%% file: main.tex
\documentclass{article}

\usepackage[final]{neurips_2020}

\usepackage[]{neurips_2020}
\input{math_commands.tex}

\usepackage[utf8]{inputenc} 
\usepackage[T1]{fontenc}    
\usepackage{hyperref}       
\usepackage{url}            
\usepackage{booktabs}       
\usepackage{amsfonts}       
\usepackage{nicefrac}       
\usepackage{microtype}      
\newcommand{\settitle}{\@maketitle}

\title{MetaPoison: Practical General-purpose Clean-label Data Poisoning}

\author{
  W. Ronny Huang\thanks{Authors contributed equally.}\\
  University of Maryland\\
  \texttt{wronnyhuang@gmail.com}\\
  \And
  Jonas Geiping\samethanks[1]\\
  University of Siegen\\
  \texttt{jonas.geiping@uni-siegen.de}\\
  \And
  Liam Fowl\\
  University of Maryland\\
  \texttt{lfowl@math.umd.edu}\\
  \AND
  Gavin Taylor\\
  United States Naval Academy\\
  \texttt{taylor@usna.edu}\\
  \And
  Tom Goldstein\\
  University of Maryland\\
  \texttt{tomg@cs.umd.edu}\\
}

\usepackage{subcaption}
\usepackage{algorithm}
\usepackage{algorithmic}
\usepackage{wrapfig}
\usepackage{amsmath}
\usepackage{amsfonts}
\usepackage{xcolor}
\usepackage{amssymb}
\usepackage{epsfig}
\usepackage[font=small]{caption}
\usepackage[export]{adjustbox}

\newcommand*\samethanks[1][\value{footnote}]{\footnotemark[#1]}

\usepackage{hyperref}
\begin{document}

\maketitle

\vspace{-18pt}
\begin{abstract}
\vspace{-3pt}

   
Data poisoning---the process by which an attacker takes control of a model by making imperceptible changes to a subset of the training data---is an emerging threat in the context of neural networks. Existing attacks for data poisoning neural networks have relied on hand-crafted heuristics, because solving the poisoning problem directly via bilevel optimization is generally thought of as intractable for deep models. We propose MetaPoison, a first-order method that approximates the bilevel problem via meta-learning and crafts poisons that fool neural networks. MetaPoison is effective: it outperforms previous clean-label poisoning methods by a large margin. MetaPoison is robust: poisoned data made for one model transfer to a variety of victim models with unknown training settings and architectures. MetaPoison is general-purpose, it works not only in fine-tuning scenarios, but also for end-to-end training from scratch, which till now hasn't been feasible for clean-label attacks with deep nets. MetaPoison can achieve arbitrary adversary goals---like using poisons of one class to make a target image don the label of another arbitrarily chosen class. Finally, MetaPoison works in the real-world. We demonstrate for the first time successful data poisoning of models trained on the black-box Google Cloud AutoML API.
\end{abstract}

\input{intro.tex}

\input{method.tex}

\input{experiments.tex}

\input{conclusion.tex}
\input{impact.tex}
\input{acknowledgments.tex}

\bibliographystyle{plainnat}
\bibliography{library}

\newpage
\input{supplementary.tex}

\end{document}


\twocolumn[
\icmltitle{MetaPoison: Robust General-purpose Clean-label Data Poisoning}



\icmlsetsymbol{equal}{*}

\begin{icmlauthorlist}
\icmlauthor{Aeiau Zzzz}{equal,to}
\icmlauthor{Bauiu C.~Yyyy}{equal,to,goo}
\icmlauthor{Cieua Vvvvv}{goo}
\icmlauthor{Iaesut Saoeu}{ed}
\icmlauthor{Fiuea Rrrr}{to}
\icmlauthor{Tateu H.~Yasehe}{ed,to,goo}
\icmlauthor{Aaoeu Iasoh}{goo}
\icmlauthor{Buiui Eueu}{ed}
\icmlauthor{Aeuia Zzzz}{ed}
\icmlauthor{Bieea C.~Yyyy}{to,goo}
\icmlauthor{Teoau Xxxx}{ed}
\icmlauthor{Eee Pppp}{ed}
\end{icmlauthorlist}

\icmlaffiliation{to}{Department of Computation, University of Torontoland, Torontoland, Canada}
\icmlaffiliation{goo}{Googol ShallowMind, New London, Michigan, USA}
\icmlaffiliation{ed}{School of Computation, University of Edenborrow, Edenborrow, United Kingdom}

\icmlcorrespondingauthor{Cieua Vvvvv}{c.vvvvv@googol.com}
\icmlcorrespondingauthor{Eee Pppp}{ep@eden.co.uk}

\icmlkeywords{Machine Learning, ICML}

\vskip 0.3in
]



\printAffiliationsAndNotice{\icmlEqualContribution} 

\begin{abstract}
   
   Data poisoning---the process by which an attacker takes control of a model by making imperceptible changes to a subset of the training data---is an emerging threat in the context of neural networks. Existing approaches for data poisoning have relied on heuristics, which limits their performance and versatility. Here, we formulate the task of crafting poisons more generally as a bi-level optimization problem, wherein the inner level corresponds to training a network on a poisoned dataset and the outer level corresponds to updating those poisons to achieve a desired behavior on the trained model. We then propose MetaPoison, a first-order method to solve this optimization quickly. MetaPoison is effective: it outperforms previous clean-label poisoning methods by a large margin under the same setting. MetaPoison is also robust: its poisons transfer to a variety of victims with unknown hyperparameters and architectures. Finally, MetaPoison is general-purpose: it enables a host of novel poisoning schemes. For instance, while previous poisoning methods only work when the victim is fine-tuning a pre-trained model, MetaPoison can poison end-to-end training from scratch with remarkable success, for example causing a target image to be misclassified 90\% of the time via manipulating just 1\% of the trained-from-scratch dataset. In addition, MetaPoison can achieve arbitrary adversary goals---like causing a chosen image to move into a chosen class.
   
   
\end{abstract}

\input{supp.tex}

%% file: math_commands.tex
\usepackage{amsmath,amsfonts,bm}

\def\eqref#1{equation~\ref{#1}}

\def\1{\bm{1}}

\def\eps{{\epsilon}}

\DeclareMathAlphabet{\mathsfit}{\encodingdefault}{\sfdefault}{m}{sl}
\SetMathAlphabet{\mathsfit}{bold}{\encodingdefault}{\sfdefault}{bx}{n}



%% file: intro.tex

\vspace{-14pt}
\section{Introduction}
\label{sec:intro}
\vspace{-4pt}

Neural networks are susceptible to a range of security vulnerabilities that compromise their real-world reliability. The bulk of work in recent years has focused on evasion attacks \cite{szegedy2013intriguing,athalye_obfuscated_2018}, where an input is slightly modified at inference time to change a model's prediction. These methods rely on access to the inputs during inference, which is not always available in practice.
Another type of attack is that of backdoor attacks \citep{turner2019cleanlabel, chen2017targeted, saha2019hidden}. Like evasion attacks, backdoor attacks require adversary access to model inputs during inference; notably backdoor ``triggers'' need to be inserted into the training data and then later into the input at inference time. Unlike evasion and backdoor attacks, \emph{data poisoning} does not require attacker control of model inputs at inference time. Here the attacker controls the model by adding manipulated images to the training set. These malicious images can be inserted into the training set by placing them on the web (social media, multimedia posting services, collaborative-editing forums, Wikipedia) and waiting for them to be scraped by dataset harvesting bots.
They can also be added to the training set by a malicious insider who is trying to avoid detection. A data corpus can also be compromised when arbitrary users may contribute data, such as face images for a recognition and re-identification system. 

\begin{wrapfigure}[19]{l}{0.5\textwidth}
    \vspace{-11pt}
    \centering
    \includegraphics[width=0.5\textwidth]{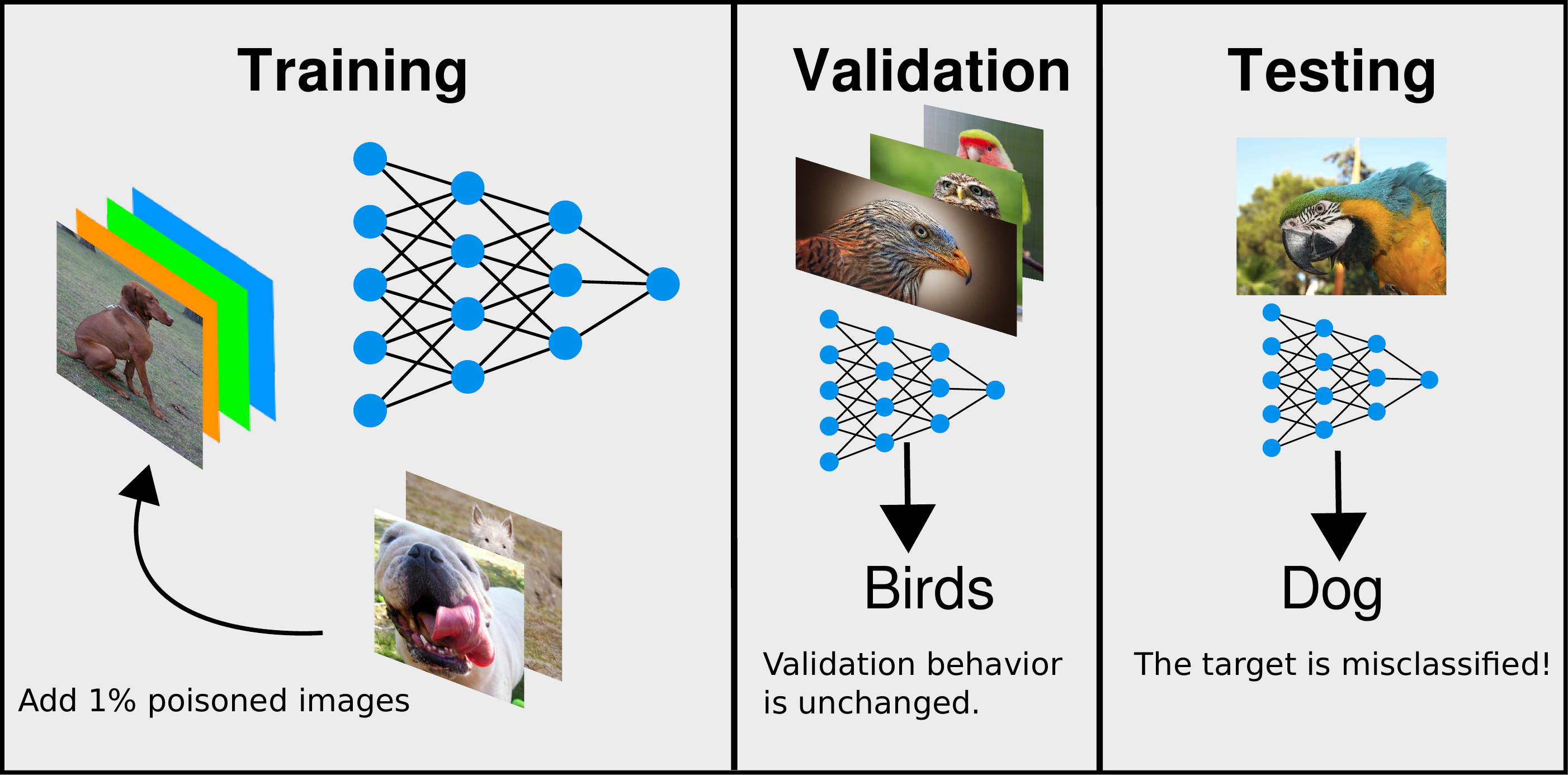}
    \vspace{-15pt}
    \caption{The attacker's goal is to classify some bird image (here: the parrot) as a dog. To do so, a small fraction of the training data is imperceptibly modified before training. The network is then trained from scratch with this modified dataset. After training, validation performance is normal (eagle, owl, lovebird). However, the minor modifications to the training set cause the (unaltered) target image (parrot) to be misclassified by the neural network as ``dog'' with high confidence.}
    \label{fig:teaser}
     \vspace{-1pt}
\end{wrapfigure}

Data poisoning attacks have been explored for classical scenarios \citep{biggio_poisoning_2012, steinhardt2017poison, burkard2017analysis} which allow both training inputs and labels to be modified.
However, it is possible to make poison perturbations imperceptible to a human observer, as they are in evasion attacks. Attacks of this type, schematic in Figure \ref{fig:teaser}, are often referred to as {\em clean-label} poisoning attacks \citep{koh2017understanding, shafahi2018poison} because poison images appear to be unmodified and labeled correctly.
The perturbed images often affect classifier behavior on a \textit{specific} target instance that comes along after a system is deployed, without affecting behavior on other inputs, making clean-label attacks insidiously hard to detect. 

Data poisoning has been posed as a bilevel optimization problem \citep{biggio_poisoning_2012, bennett_bilevel_2008}, with the higher-level objective of minimizing adversarial loss on target images depending on the lower-level objective of minimizing training loss on poisoned data. This formulation is used to generate poisoned data for SVMs \citep{biggio_poisoning_2012}, logistic regression \citep{demontis_why_2019} or linear regression \citep{jagielski_manipulating_2018-1}. However, solving the bilevel optimization problem requires differentiation w.r.t to the minimizer of the lower-level problem. This is intractable for deep neural networks, due to their inherent complexity and reliance on large datasets. \citet{munoz-gonzalez_towards_2017} and \citet{ mei_using_2015} apply back-gradient optimization to differentiate by unrolling  effectively the entire training objective, yet while this attack compromises simple learning models, it does not work for deep neural networks, leading \citet{munoz-gonzalez_towards_2017} to conclude neural networks to ``be more resilient against [...] poisoning attacks'', compared to other learning algorithms.

Due to these limitations of classical strategies, heuristic approaches, such as Feature Collision (FC), are currently the dominant approach to clean-label poisoning \citep{shafahi2018poison, zhu2019transferable}. Perturbations are used to modify a training image (e.g., a tree) so that its feature representation is nearly identical to that of a chosen target image (e.g., a stop sign). After the victim fine tunes their model on the poisoned image, the model cannot distinguish between the poison and target image, causing it to misclassify the stop sign as a tree.
%
FC is a heuristic with limited applicability;  the attacker must have knowledge of the feature extractor being used, and the feature extractor cannot substantially change after the poison is introduced.  For this reason, FC attacks only work on fine-tuning and transfer learning pipelines, and fail when the victim trains their model from scratch.
Also, FC is not general-purpose---an attacker could have objectives beyond causing a single target instance to be misclassified with the label of the poison.

Our contributions are fivefold. First, we re-evaluate bilevel optimization for data poisoning of deep neural networks and discover a key algorithm, henceforth called MetaPoison, that allows for an effective approximation of the bilevel objective. Second, in contrast to previous approaches based on bilevel optimization, we outperform FC methods by a large margin in the established setting where a victim fine-tunes a pre-trained model. Third, we demonstrate, for the first time, successful clean-label poisoning in the challenging context where the victim trains deep neural nets \textit{from scratch} using random initializations. Fourth, we show that MetaPoison can enable alternative, never-before-demonstrated poisoning schemes. Fifth, we verify MetaPoison's practicality in the real world by successfully poisoning models on the black-box Google Cloud AutoML API platform.

End-to-end code as well as pre-crafted poisons are available at \url{https://www.github.com/wronnyhuang/metapoison}. 
We encourage the reader to download, train, and evaluate our poisoned CIFAR-10 dataset on their own CIFAR-10 training pipeline to verify MetaPoison's effectiveness. 
Note finally that MetaPoison can also be used for non-nefarious purposes, such as copyright enforcement. For example, it can ``watermark'' copyrighted data with diverse, undetectable perturbations. The model can then be queried with the target (known only to copyright holder) to determine whether the copyrighted data was used to train the model.

%% file: method.tex
\vspace{-8pt}
\section{Method}
\vspace{-4pt}
\subsection{Poisoning as constrained bilevel optimization}
\vspace{-4pt}

Suppose an attacker wishes to force an unaltered target image $x_t$ of their choice to be assigned an incorrect, \textit{adversarial} label $y_\text{adv}$ by the victim model. The attacker can add $n$ poison images $X_p\in[0, 255]^{n\times m}$, where $m$ is the number of pixels, to the victim's clean training set $X_c$.
The optimal poison images $X_p^*$ can be written as the solution to the following optimization problem:

\vspace{-9pt}
\begin{equation}
    X_p^* = \underset{X_p}{\mathrm{argmin}}\; \mathcal{L}_\textrm{adv}(x_t, y_\text{adv} ; \theta^*(X_p)),
    \label{bilevel}
\end{equation}

\vspace{-9pt}
where in general $\mathcal{L}(x, y; \theta)$ is a loss function measuring how accurately a model with weights $\theta$ assigns label $y$ to input $x$. For $\mathcal{L}_\textrm{adv}$ we use the \citet{carlini2017towards} $f_6$ function and call it the \textit{adversarial loss}.
$\theta^*(X_p)$ are the network weights found by training on the poisoned training data $X_c \cup X_p$, which contain the poison images $X_p$ mixed in with mostly clean data $X_c\in[0, 255]^{N\times m}$, where $N\gg n$. Note that (\ref{bilevel}) is a bi-level optimization problem \citep{bard_practical_2013} -- the minimization for $X_p$ involves the weights $\theta^*(X_p)$, which are themselves the minimizer of the training problem,

\vspace{-9pt}
\begin{equation}
    \theta^*(X_p) = \underset{\theta}{\mathrm{argmin}}\; \mathcal{L}_\textrm{train}(X_c \cup X_p, Y ;\theta),
    \label{eq:lower_level}
\end{equation}

\vspace{-9pt}
where $\mathcal{L}_\textrm{train}$ is the standard cross entropy loss, and $Y\in\mathbb{Z}^{N + n}$ contains the correct labels of the clean and poison images. Thus, (\ref{bilevel}) and (\ref{eq:lower_level}) together elucidate the high level formulation for crafting poison images: find $X_p$ such that the \textit{adversarial loss} $\mathcal{L}_\textrm{adv}(x_t, y_\textrm{adv}; \theta^*(X_p))$ is minimized after training.

For the attack to be inconspicuous, each poison example $x_p$ should be constrained to ``look similar'' to a natural base example $x$. A number of perceptually aligned perturbation models have been proposed  \citep{engstrom2019exploring, wong2019wasserstein,ghiasi2020breaking}. We chose the ReColorAdv perturbation function of \citet{laidlaw2019functional}, which applies a function $f_g$, with parameters $g$, and an additive perturbation map $\delta$, resulting in a poison image $x_p = f_g(x) + \delta$. The function $f_g(x)$ is a pixel-wise color remapping $f_g: \mathcal{C}\rightarrow\mathcal{C}$ where $\mathcal{C}$ is the 3-dimensional LUV color space. 
To ensure that the perturbation is minimal, $f_g$ can be bounded such that for every pixel $x_i$, $\|f_g(x_i) - x_i\|_\infty < \eps_c$, and $\delta$ can be bounded such that $\|\delta\|_\infty < \eps$. We use the standard additive bound of $\eps = 8$ and a tighter-than-standard color bound of $\eps_c = 0.04$ to further obscure the perturbation (\citet{laidlaw2019functional} used $\eps_c = 0.06$). To enforce these bounds, we optimize for $X_p$ with PGD \citep{madry2017towards}, projecting the outer-parameters $g$ and $\delta$ back to their respective $\eps_c$ and $\eps$ balls after every gradient step. Example poisons along with their clean counterparts used in this work are shown later in Figure \ref{fig:central} (top left).

\vspace{-4pt}
\subsection{Strategy for crafting effective poisoning examples}
\label{sec:mainalgo}
\vspace{-4pt}

Minimizing the full bi-level objective in (\ref{bilevel})-(\ref{eq:lower_level}) is intractable.
We can, however, approximate the inner objective (the training pipeline) by training only $K$ SGD steps for each outer objective evaluation.
This allows us to ``look ahead'' in training and view how perturbations to poisons \textit{now} will impact the adversarial loss $K$ steps \textit{later}. For example, the process of unrolling two inner-level SGD steps to compute an outer-level update on the poisons would be
\begin{align}
    \vspace{-5pt}
    \theta_1 &= \theta_{0} - \alpha\nabla_\theta\mathcal{L}_\textrm{train}(X_c \cup X_p, Y; \theta_{0}) \nonumber \\
    \theta_2 &= \theta_1 - \alpha\nabla_\theta\mathcal{L}_\textrm{train}(X_c \cup X_p, Y; \theta_1) \nonumber \\
    X_p^{i+1} &= X_p^{i} - \beta\nabla_{X_p}\mathcal{L}_\textrm{adv}(x_t, y_\textrm{adv}; \theta_2), \label{eq:unroll3}
\end{align}

where $\alpha$ and $\beta$ are the learning rate and crafting rate, respectively. 
$K$-step methods have been found to have exponentially decreasing approximation error \citep{shaban2019truncated} and generalization benefits \citep{franceschi2018bilevel}.

\begin{figure}[t!]
    \vspace{-19pt}
    \centering
    \begin{subfigure}[t]{0.4\textwidth}
        \centering
        \includegraphics[width=\textwidth]{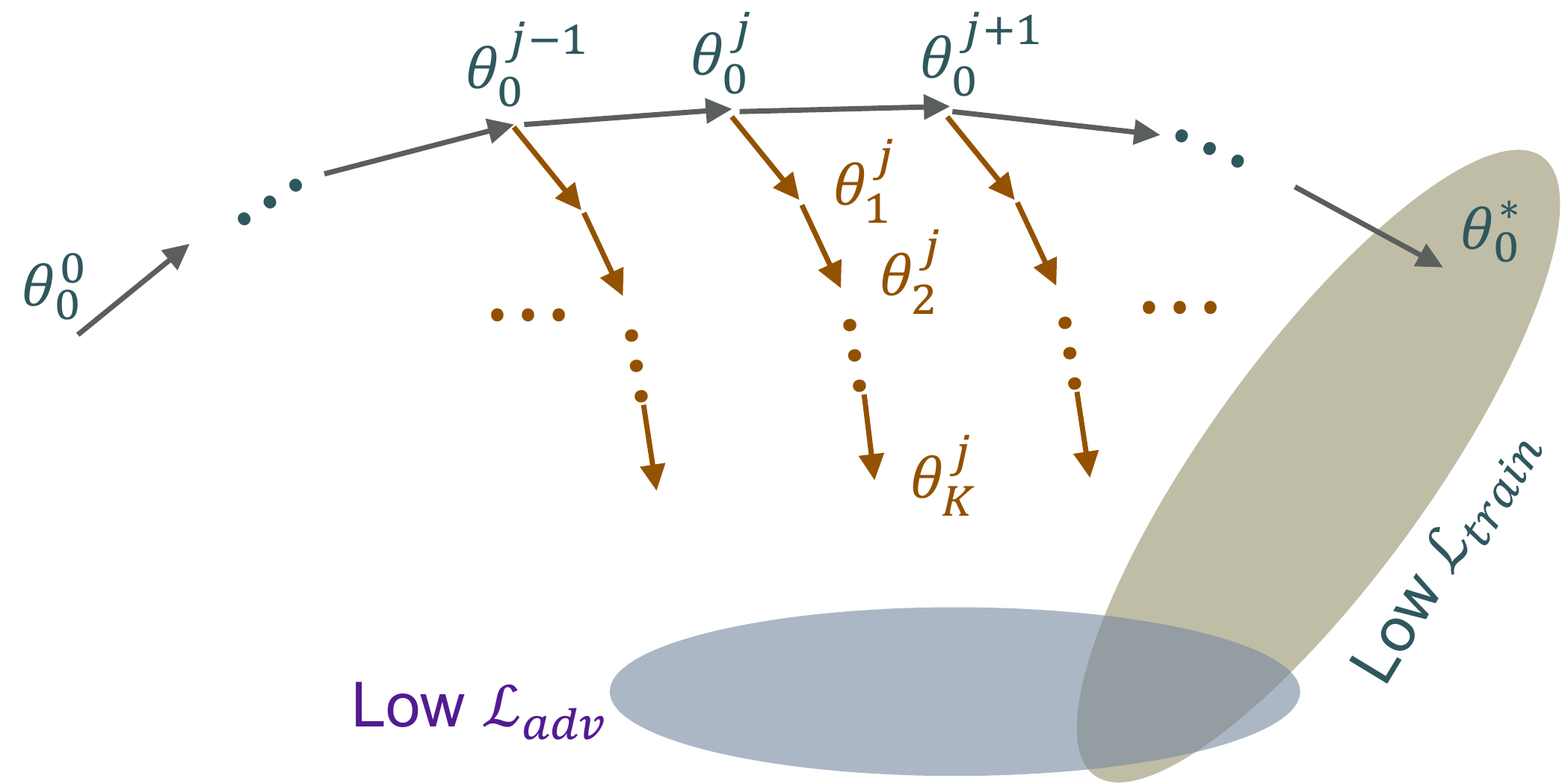}
    \end{subfigure}
    \quad\quad
    \begin{subfigure}[t]{0.4\textwidth}
        \centering
        \includegraphics[width=\textwidth]{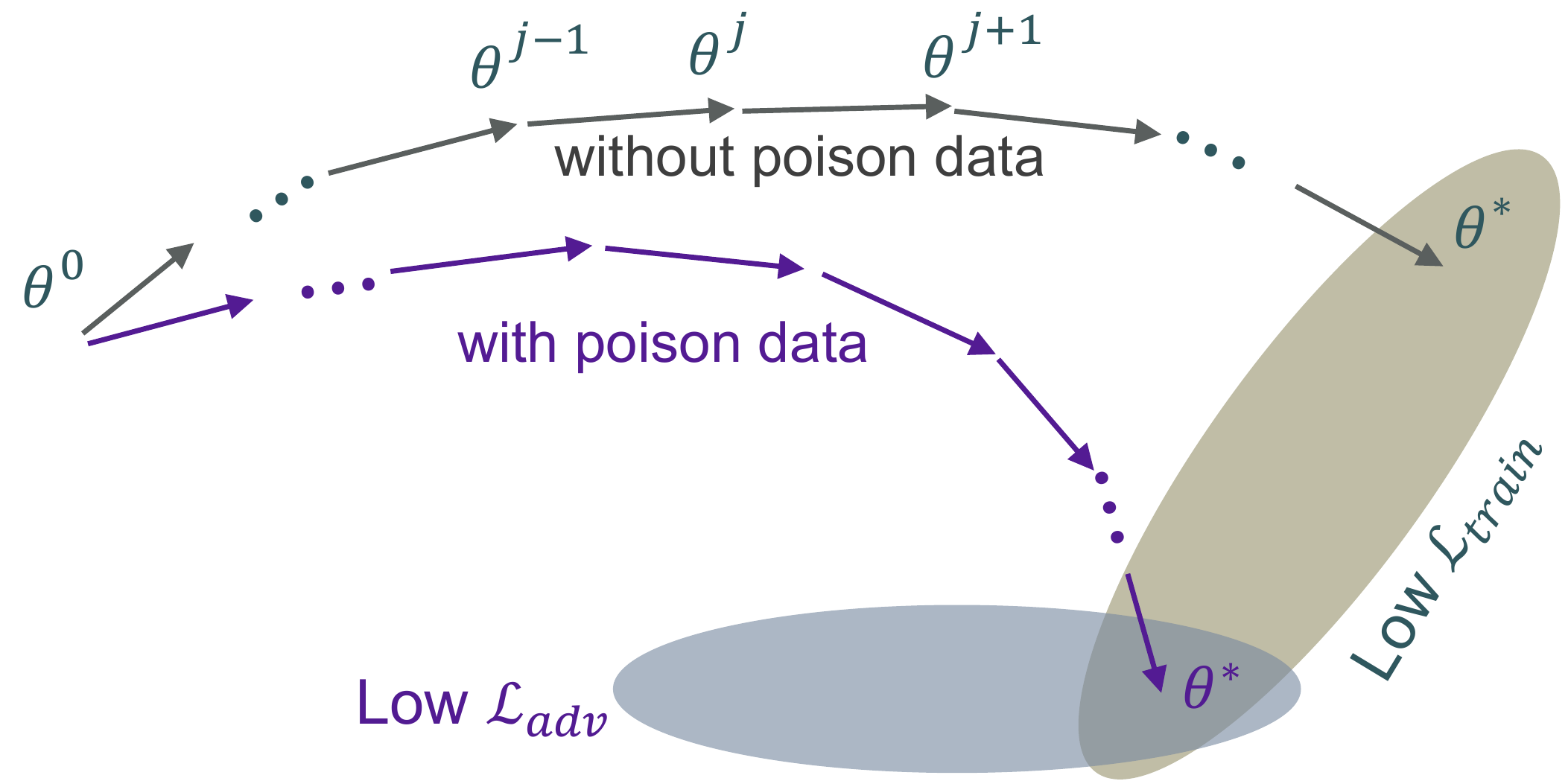}
    \end{subfigure}
    \small
    \vspace{-4pt}
    \caption{MetaPoison in weight space. Gray arrows denote normal training trajectory with weights ${\theta_0^j}$ at the $j$-th step. (Left) During the poison crafting stage, the computation graph consisting of the training pipeline is unrolled by $K$ SGD steps forward in order to compute the perturbation to the poisons $\nabla_{X_p}\mathcal{L}_\textrm{adv}$, starting from various points along the trajectory. Optimally, those poisons will steer weights (brown arrows) toward regions of low $\mathcal{L}_\textrm{adv}$ regardless of which training step $\theta_0^j$ the poisons are inserted into. (Right) When the victim trains on the poisoned data (purple arrows), the weight trajectory is collectively and implicitly steered to regions of low $\mathcal{L}_\textrm{adv}$ whilst the learner explicitly drives the weights to regions of low $\mathcal{L}_\textrm{train}$.}
    \label{fig:method}
    \vspace{-13pt}
\end{figure}

Poisons optimized this way should cause the adversarial loss $\mathcal{L}_\textrm{adv}$ to drop after $K$ additional SGD steps. Ideally, this should happen \textit{regardless} of where the poisons are inserted along the network trajectory, as illustrated in Figure \ref{fig:method} (left). Our approach, discussed in the next two paragraphs, encourages the poisons to have this property.
When inserted into the training set of a victim model, the poisons should implicitly ``steer'' the weights toward regions of low $\mathcal{L}_\textrm{adv}$ whilst the learner drives the weights toward low training loss $\mathcal{L}_\textrm{train}$. When poisoning is successful, the victim should end up with a weight vector that achieves both low $\mathcal{L}_\textrm{adv}$ and $\mathcal{L}_\textrm{train}$ despite having only explicitly trained for low $\mathcal{L}_\textrm{train}$, as shown in Figure \ref{fig:method} (right). 

\begin{wrapfigure}[28]{r}{0.54\textwidth}
\vspace{-12pt}
\begin{minipage}{0.54\textwidth}
\begin{algorithm}[H]
\caption{Craft poison examples via MetaPoison}
\label{alg:metapoison}
\small
\begin{algorithmic}[1]
    \STATE {\bfseries Input} Training set of images and labels $(X, Y)$ of size $N$, target image $x_t$, adversarial class $y_\mathrm{adv}$, $\eps$ and $\eps_c$ thresholds, $n \ll N$ subset of images to be poisoned, $T$ range of training epochs, $M$ randomly initialized models.
    \STATE {\bfseries Begin}
    \STATE Stagger the $M$ models, training the $m$th model weights $\theta_m$ up to  $\left \lfloor{mT/M}\right \rfloor$ epochs
    \STATE Select $n$ images from the training set to be poisoned, denoted by $X_p$. Remaining clean images denoted $X_c$
    \STATE For $i = 1, \dots, C$ crafting steps:
    \STATE \quad For $m=1,\dots, M$ models:
    \STATE \quad \quad Copy $\tilde\theta = \theta_m$
    \STATE \quad \quad For $k=1,\dots, K$ unroll steps$^a$:
    \STATE \quad \quad \quad $\tilde\theta = \tilde\theta - \alpha \nabla_{\tilde\theta} \mathcal{L}_\mathrm{train}(X_c \cup X_p, Y; \tilde\theta)$
    \STATE \quad \quad Store adversarial loss $\mathcal{L}_m = \mathcal{L}_\mathrm{adv}(x_t, y_\textrm{adv}; \tilde\theta)$
    \STATE \quad \quad Advance epoch $\theta_m\!=\!\theta_m\!-\!\alpha \nabla_{\theta_m}\mathcal{L}_\mathrm{train}(X, Y; \theta_m)$
    \STATE \quad \quad If $\theta_m$ is at epoch $T + 1$:
    \STATE \quad \quad \quad Reset $\theta_m$ to epoch 0 and reinitialize
    \STATE \quad Average adversarial losses $\mathcal{L}_\mathrm{adv} = \sum_{m=1}^M \mathcal{L}_m/M$
    \STATE \quad Compute $\nabla_{X_p} \mathcal{L}_\mathrm{adv}$
    \STATE \quad Update $X_p$ using Adam and project onto $\eps, \eps_c$ ball
    \STATE {\bfseries Return} $X_p$ \\
    \footnotesize $^a$For brevity, we write as if unrolled SGD steps are taken using the full dataset. In practice they are taken on minibatches and repeated until the full dataset is flushed once through. The two are effectively equivalent.
\end{algorithmic}
\end{algorithm}
\end{minipage}
\end{wrapfigure}

The idea of unrolling the training pipeline to solve an outer optimization problem has been successfully applied to meta-learning \citep{finn2017model}, hyperparameter search \citep{maclaurin2015gradient, domke2012generic}, architecture search \citep{liu2018darts}, and poisoning of shallow models \citep{munoz-gonzalez_towards_2017}.
However, unique challenges arise when using this method for robust data poisoning of deep models. First, the training process depends on weight initialization and minibatching order, which are determined at random and unknown to the attacker.  This is in contrast to meta-learning, hyperparameter search, and architecture search, where the same agent has purview into both the inner (training their \textit{own} networks) and outer processes. Second, we find that using a single surrogate network to craft poisons causes those poisons to overfit to the weights of that network at \textit{that} epoch, while failing to steer new, randomly initialized weights toward low $\mathcal{L}_\textrm{adv}$. In other words, data poisoning demands less a solution that perfectly solves the bilevel problem (\ref{bilevel}) for one model than one that generalizes to new networks with different initializations and at different epochs.

We address the problem of generalization via \textit{ensembling} and \textit{network re-initialization}. Poisons are crafted using an ensemble of partially trained surrogate models \textit{staggered by epoch}. The update to the poisons has the form, 
\vspace{-6pt}
\begin{equation}
    X_p^{i+1} = X_p^i - \frac{\beta}{N_\textrm{epoch}} \nabla_{X_p} \sum_{j=0}^{N_\textrm{epoch}} \mathcal{L}_\textrm{adv}\Big|_{\theta^j}, \label{eq:ensemble}
\end{equation}
where $\mathcal{L}_\textrm{adv}\vert_{\theta^j}$ is the adversarial loss after a few look-ahead SGD steps on the poisoned dataset starting from weights $\theta^j$ from the $j$-th epoch. The update gradient, $\nabla_{X_p}\mathcal{L}_\textrm{adv}$, was explicitly written out in (\ref{eq:unroll3}) for one model, where the starting weight here $\theta^j$ here corresponds to $\theta_0$ in (\ref{eq:unroll3}). The summation in (\ref{eq:ensemble}) averages the adversarial loss over the ensemble, where each model in the ensemble is at a different epoch denoted by $\theta^j$. This forces the poisons to be effective when inserted into a minibatch at any stage of training. Between each poison update, the set of weight vectors $\{\theta^j\}$ are vanilla-trained for a single epoch; once a model has trained for a sentinel number of epochs, it is randomly re-initialized back to epoch 0. This forces the poisons to adapt to diverse network initializations.
%
%
The entire process is outlined in Algorithm \ref{alg:metapoison}. 

Based on our experimental settings (\S\ref{sec:expt}), MetaPoison takes 2 (unrolling steps) $\times$ 2 (backprop thru unrolled steps) $\times$ 60 (outer steps) $\times$ 24 (ensemble size) $=$ 5760 forward+backward propagations per poison. In contrast \citet{shafahi2018poison} reports 12000 forward+backward props. Thus MetaPoison has similar cost if we discount the one-time pretraining of the surrogate models. Crafting 500 poisons for 60 steps on CIFAR-10 takes about 6 GPU-hours and can be shortened to 5 GPU-hours by loading pretrained surrogate model checkpoints.


It is worth discussing why this strategy of crafting poisons is effective. In contrast to previous works we significantly alter the gradient estimation for the inner-level objective. First, we make $K$ (the number of unrolled steps) small---we choose $K=2$ for all examples in this work, whereas $K$ is chosen within $60-200$ for deep networks in \citet{munoz-gonzalez_towards_2017} and whereas the entire algorithm is unrolled in \citet{maclaurin2015gradient, domke2012generic,mei_using_2015}, corresponding to $K\approx 10^5$ in our setting. This choice is supported by \citet{shaban2019truncated}, which proved that under mild conditions, the approximation error of few $K$ step evaluations decreases exponentially, and by \citet{maclaurin2015gradient}, which discussed that due to the ill-posedness of the gradient operator, even for convex problems, the numerical error increases with each step. Both taken together imply that most of the gradient can be well approximated within the first steps, whereas later steps, especially with the limited precision, possibly distort the gradient. Another consideration is generalization. In comparison to (\ref{bilevel}), the full bilevel objective for an unknown victim model trained from-scratch contains two additional sources of randomness, the \textit{random initialization} of the network and the \textit{random stochastic gradient} descent (SGD) direction over prior steps. So, for practical success, we need to reliably estimate gradients of this probabilistic objective. Intuitively, and shown in \cite[Sec. 5.1]{franceschi2018bilevel}, the exact computation of the bilevel gradient for a \textit{single} arbitrary initialization and SGD step leads to overfitting, yet keeping $K$ small acts as an implicit regularizer for generalization. Likewise, both reinitializing the staggered models and ensembling a variety of such models are key factors that allow for a reliable estimate of the full train-from-scratch objective, which we can view as expectation value over model initialization and SGD paths. The appendix substantiates via ablation studies the importance or viability of small $K$ (\S\ref{sec:unrollsteps}), ensembling (\S\ref{sec:ablationensembling}), and network reinitialization (\S\ref{sec:ablationreinit}).

%% file: experiments.tex
\vspace{-5pt}
\section{Experiments}
\vspace{-5pt}
\label{sec:expt}

Our experiments on CIFAR-10 consist of two stages: poison crafting and victim evaluation. In the first stage, we craft poisons on surrogate models and save them for evaluation. In the second stage, we insert the poisons into the victim dataset, train the victim model from scratch on this dataset, and report the attack success and validation accuracy. We declare an attack successful only if the target instance $x_t$ is classified as the adversarial class $y_\mathrm{adv}$; it doesn't count if the target is classified into any other class, incorrect or not. The attack success rate is defined as the number of successes over the number of attacks attempted. Unless stated otherwise, our experimental settings are as follows. The first $n$ examples in the poisons' class are used as the base images in the poison set $X_p$ and are perturbed, while the remaining images in CIFAR-10 are used as the clean set $X_c$ and are untouched. The target image is taken from the CIFAR-10 test set. We perform 60 outer steps when crafting poisons using the Adam optimizer with an initial learning rate of 200. We decay the outer learning rate (i.e. crafting rate) by 10x every 20 steps. 
Each inner learner is unrolled by $K=2$ SGD steps. 
An ensemble of 24 inner models is used, with model $i$ trained until the $i$-th epoch. A batchsize of 125 and learning rate of 0.1 are used. We leave weight decay and data augmentation off by default, but analyze performance with them on in \S\ref{sec:robustness}. By default, we use the same 6-layer ConvNet architecture with batch normalization as \citet{finn2017model}, henceforth called ConvNetBN, but other architectures are demonstrated throughout the paper too. Outside of \S\ref{sec:robustness}, the same hyperparameters and architectures are used for victim evaluation. We train each victim to 200 epochs, decaying the learning rate by 10x at epochs 100 and 150.
The appendix contains ablation studies against the number of outer steps (\S\ref{sec:craftsteps}), $K$ (\S\ref{sec:unrollsteps}), perturbation (both $\epsilon$ and $\epsilon_c$) magnitude (\S\ref{sec:pertmag}), poison-target class pair (\S\ref{sec:allpairs}), and target image ID (\S\ref{sec:targids}).

\subsection{Comparison to previous work}
\label{sec:shafahi}
Previous works on clean-label poisoning from \citet{koh2017understanding}, \citet{shafahi2018poison}, and \citet{zhu2019transferable} attack models that are pre-trained on a clean/standard dataset and then fine-tuned on a poisoned dataset. We compare MetaPoison to \citet{shafahi2018poison}, who crafted poisons using feature collisions in a white-box setting where the attacker has knowledge of the pretrained CIFAR-10 AlexNet-like classifier weights. They assume the victim fine-tunes using the entire CIFAR-10 dataset.
\begin{wrapfigure}[32]{r}{.46\textwidth}
    \small
    \vspace{-4pt}
    \centering
    \includegraphics[width=.45\textwidth]{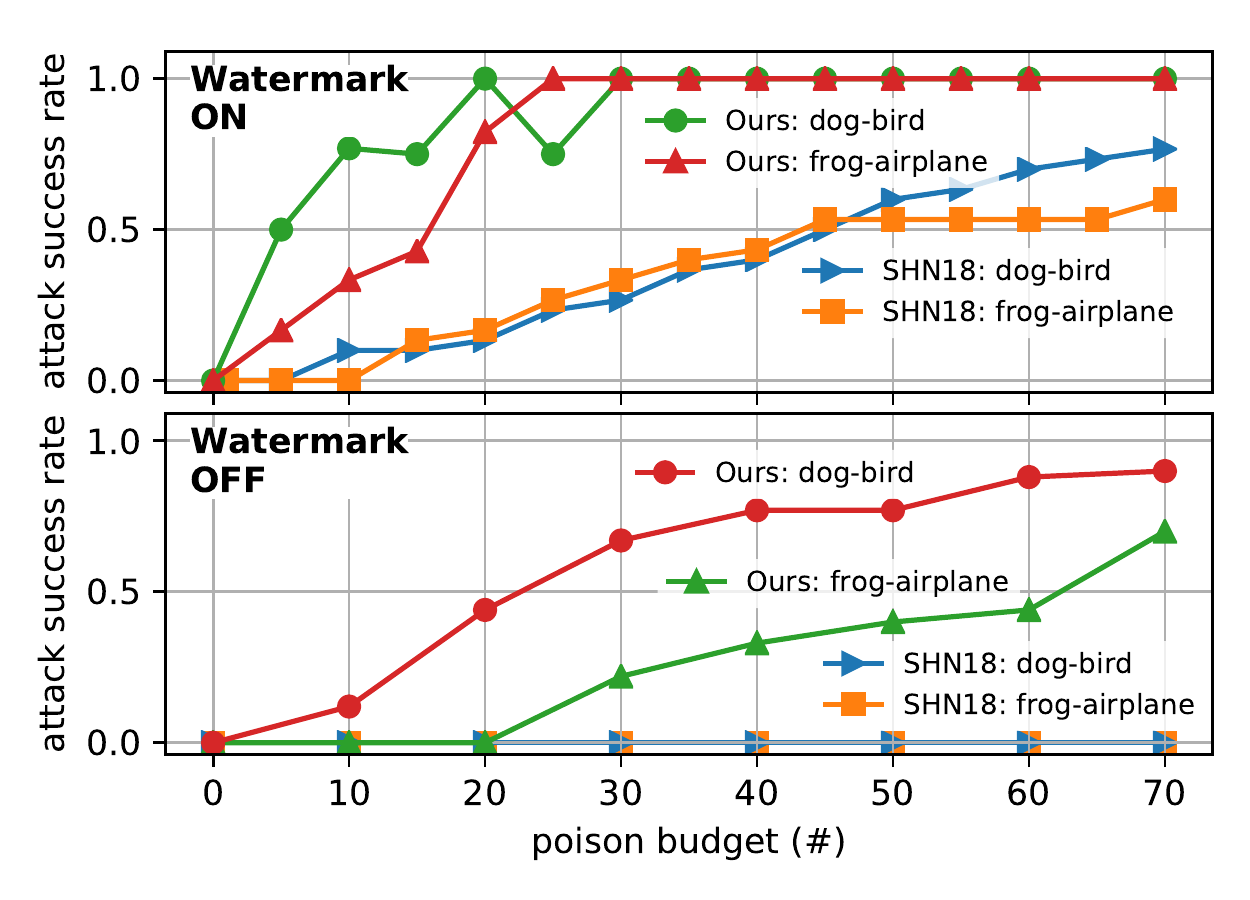} \\
    \vspace{-4pt}
    \includegraphics[width=.42\textwidth]{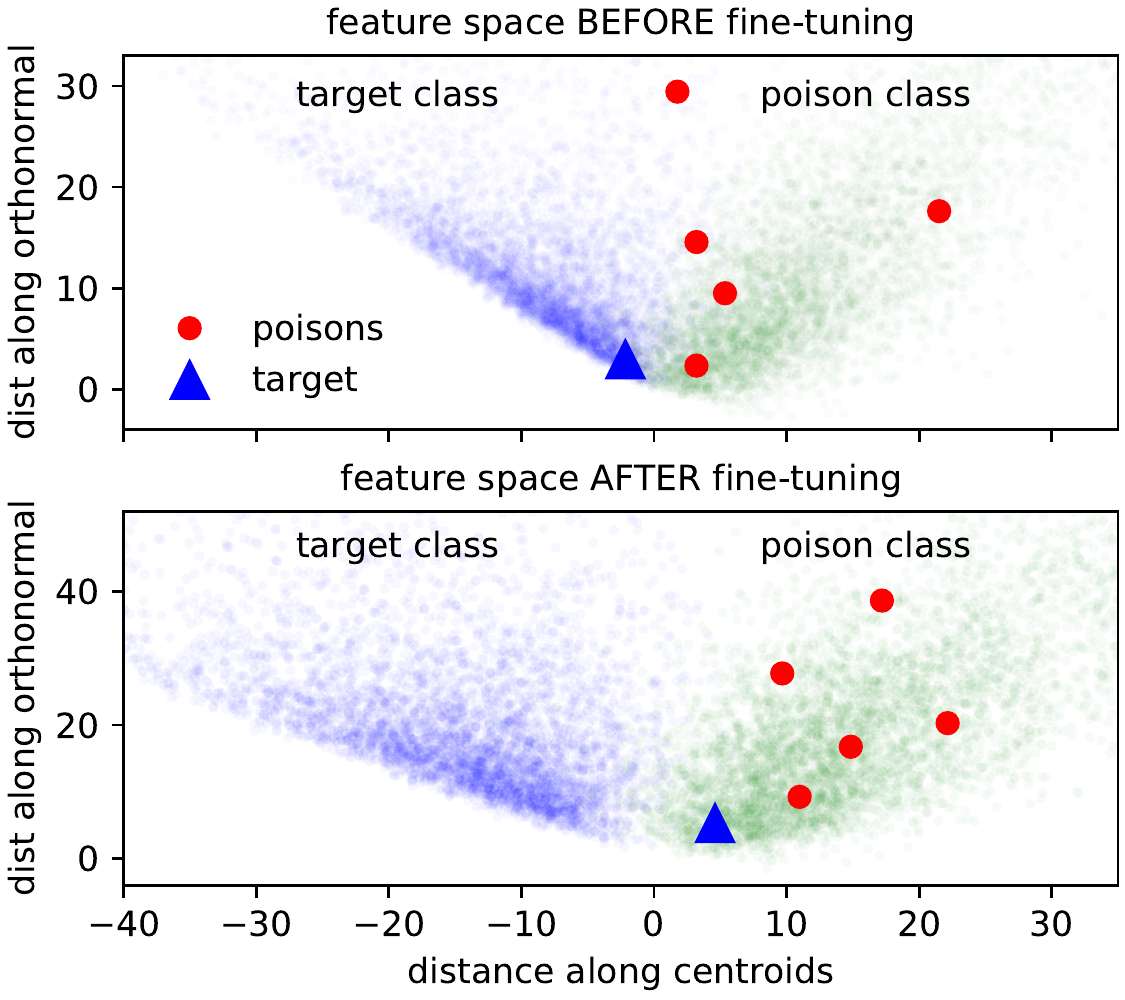}
    \vspace{-5pt}
    \caption{Comparison with \citet{shafahi2018poison} (SHN18) under the same fine-tuning conditions. (Top) Success rates for a watermark-trick opacity of 30\% or 0\%. (Bottom) Penultimate-layer feature representation visualization of the target and poison class examples before and after fine-tuning on the poisoned dataset.}
    \label{fig:shafahi-compare}
    \vspace{-10pt}
\end{wrapfigure}
Critical to their success was the ``watermark trick'': they superimpose a 30\% opacity watermark of the target image onto every poison image before crafting applying their additive perturbation.  
For evaluation, \citet{shafahi2018poison} compared two poison-target class pairs, frog-airplane and dog-bird, and ran poisoning attacks on 30 randomly selected target instances for each class pair.
They also varied the number of poisons. We replicate this scenario as closely as possible using poisons crafted via MetaPoison.
Since the perturbation model in \citet{shafahi2018poison} was additive only (no ReColorAdv), we set $\eps_c=0$ in MetaPoison.
To apply MetaPoison in the fine-tuning setting, we first pretrain a network to 100 epochs and use this fixed network to initialize weights when crafting poisons or running victim evaluations. 
Our comparison results are presented in Figure \ref{fig:shafahi-compare} (top). Notably, 100\% attack success is reached at about 25 poisons out of 50000 total training examples, or a poison budget of only 0.05\%. In general, MetaPoison achieves much higher success rates at much lower poison budgets as compared to the previous method, showcasing the strength of its poisons to alter victim behavior in the case of fine-tuning. Furthermore, MetaPoison achieves success even without the watermark trick while Shafahi et al.'s method fails, consistent with their reported ablation study.

The fine-tuning scenario also provides a venue to look closer into the mechanics of the attack. In the feature collision attack \citep{shafahi2018poison}, the poisons are all crafted to share the same feature representation as that of the target in the penultimate layer of the network. When the features in the penultimate layer are visualized\footnote{Like \citet{shafahi2018poison}, we project the representations along the line connecting centroids of the two classes (x-axis) and along the orthogonal component of the classification-layer parameter vector (y-axis). This projection ensures that we are able to see activity at the boundaries between these two classes.}, the poisons are overlapped, or collided, with the target (Figure 3b in Shafahi et al.). We perform the same visualization in Figure \ref{fig:shafahi-compare} (bottom) for a successful attack with 5 poisons using MetaPoison. Intriguingly, our poisons do \textit{not} collide with the target, implying that they employ some other mechanism to achieve the same objective. They do not even reside in the target class distribution, which may render neighborhood conformity tests such as \citet{papernot2018dknn, gupta2019strong} less effective as a defense.
Figure \ref{fig:shafahi-compare} (bottom) also shows the feature representations after fine-tuning. The target switches sides of the class boundary, and dons the incorrect poison label. 
These visualizations show that MetaPoisons cause feature extraction layers to push the target in the direction of the poison class without relying on feature collision-based mechanics.
Indeed, the poisoning mechanisms of MetaPoison are \textit{learned} rather than hand-crafted; like adversarial examples, they likely do not lend themselves to an easy human interpretation, making them difficult to detect.
Appendix \S\ref{sec:featvizapp} contains analogous feature visualizations for poisoning in the train-from-scratch context, which we discuss next.

\subsection{Victim training from scratch}
\label{sec:fromscratch}

Usually fine-tuning datasets tend to be small, domain-specific, and well-curated; from a practical perspective, it may be harder to inject poisons into them. On the other hand, large datasets on which models are (pre)trained from scratch are often scraped from the web, making them easier to poison. Thus, a general-purpose poisoning method that works on models trained from scratch would be far more viable. Yet \textit{no} prior clean-label poisoning attack has been demonstrated against networks trained from scratch. This is because existing feature collision-based poisoning \citep{shafahi2018poison, zhu2019transferable} requires a pre-existing feature extractor on which to craft a feature collision. 

\begin{figure}[t!]
    \small
    \centering
    \vspace{-1pt}
    \includegraphics[width=0.55\textwidth]{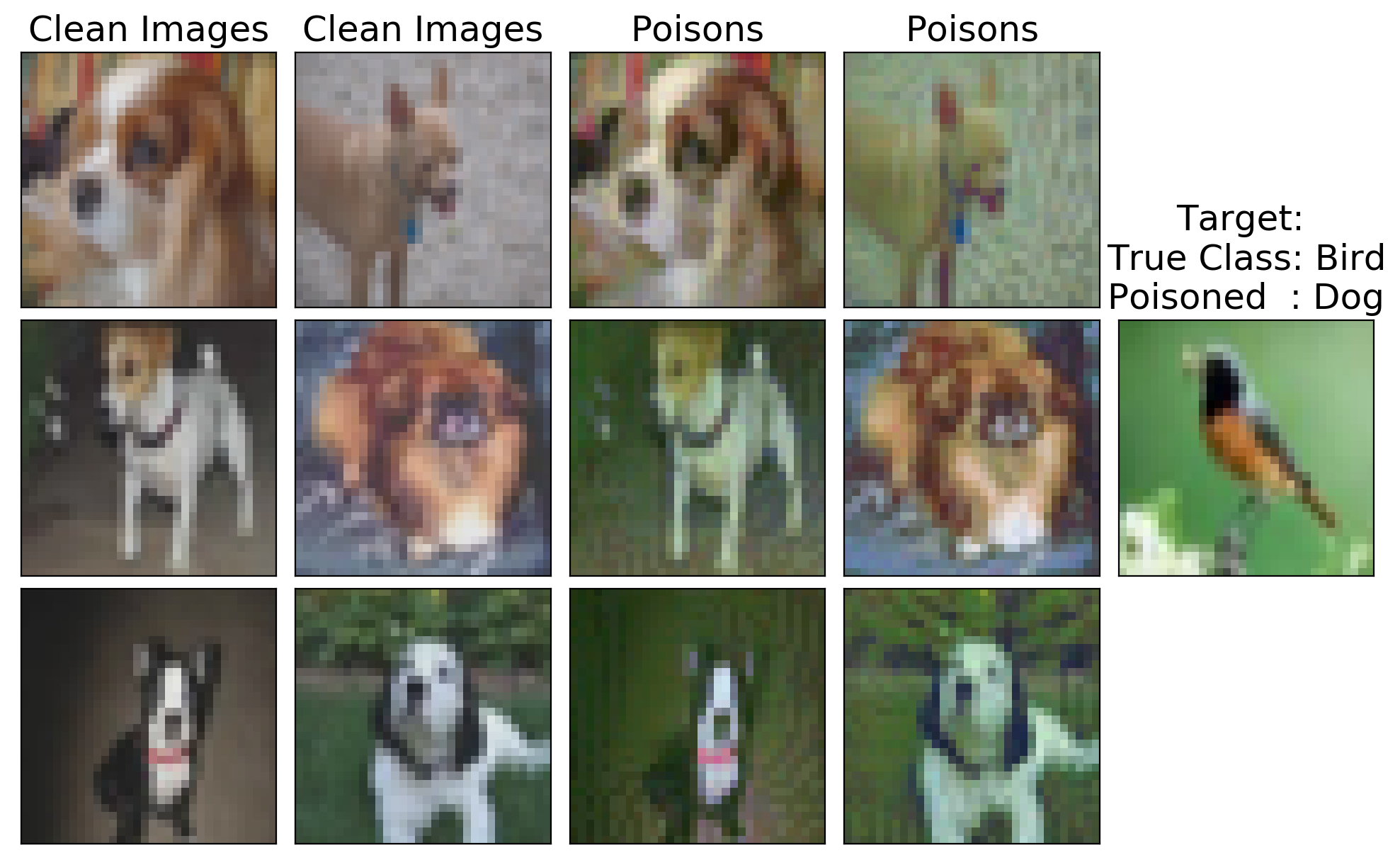} 
    \includegraphics[width=0.44\textwidth]{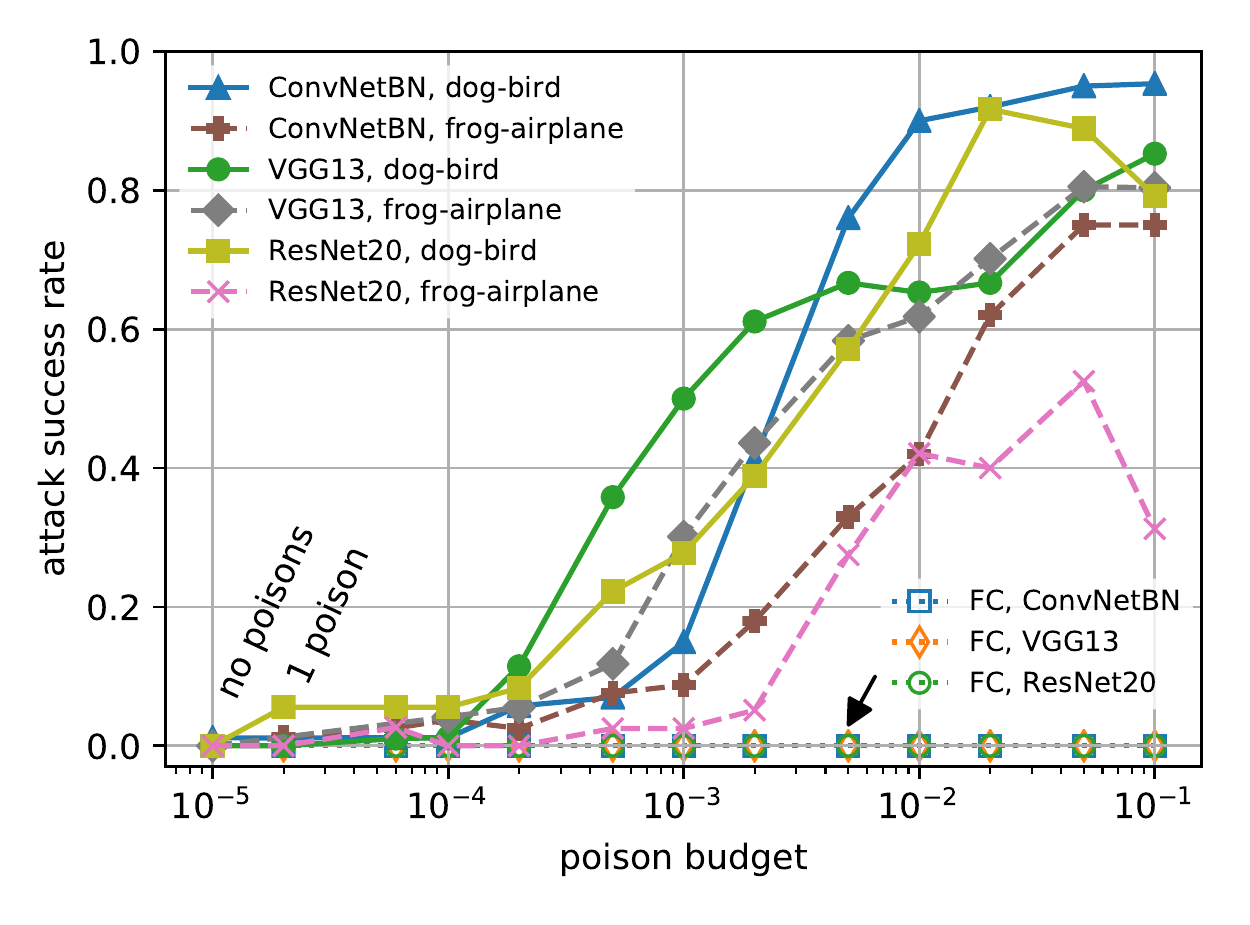} \\
    \includegraphics[width=0.55\textwidth]{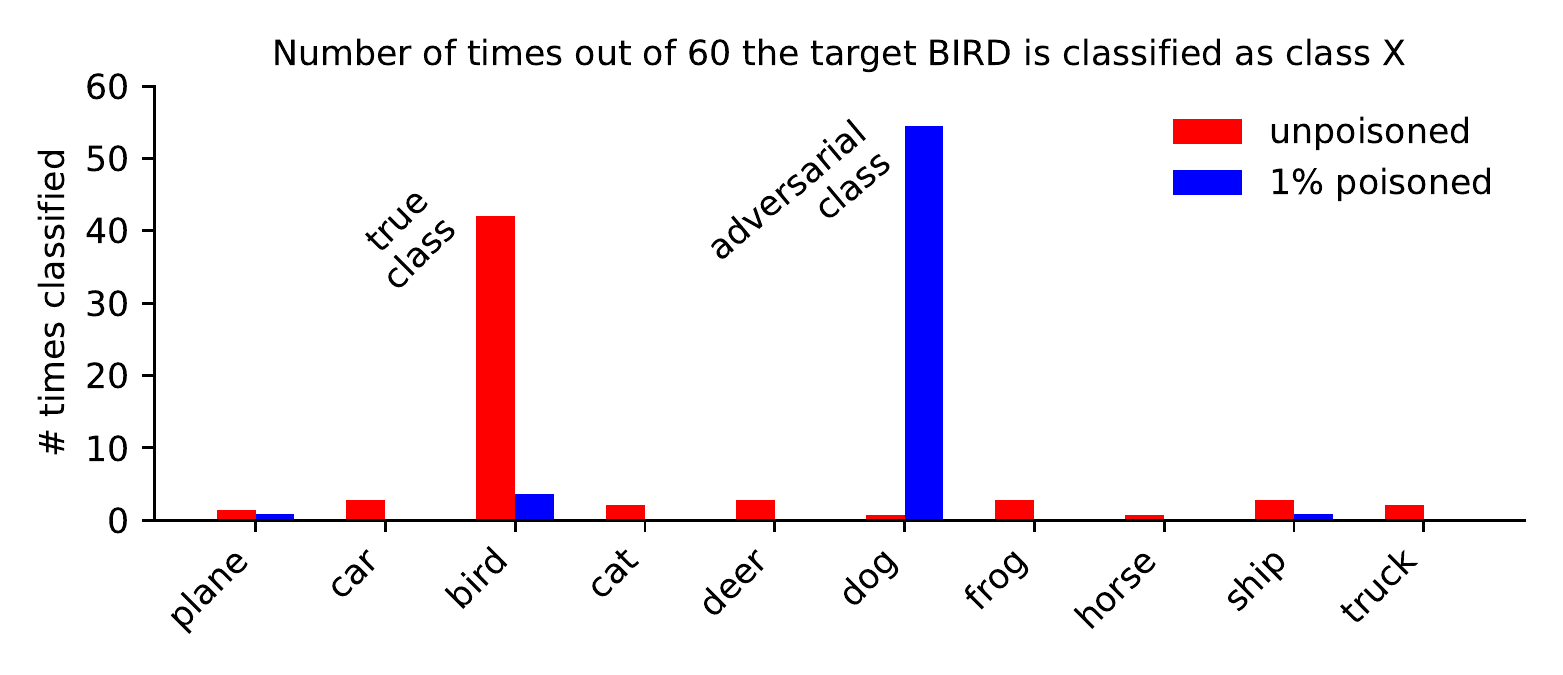}
    \includegraphics[width=0.44\textwidth]{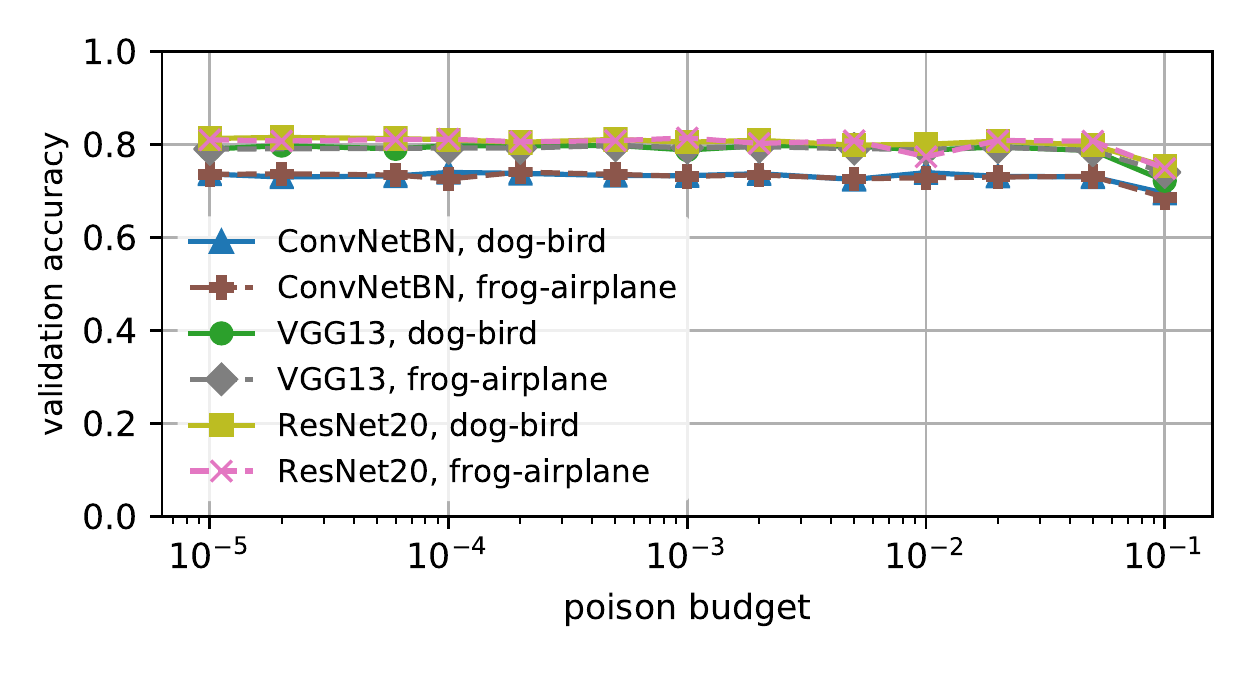}
    \vspace{-18pt}
    \caption{Poisoning end-to-end training from scratch. (Top left) Examples of poisoned training data. (Bottom left) Tally of the classes into which target birds are classified over 60 victim models on ConvNetBN. 6 models are trained with different random seeds for each of 10 target birds, totaling 60 victim models. (Top right) Attack success rate vs poison budget for different architectures and poison-target class pairs. (Bottom right) Validation accuracy of poisoned models.}
    \label{fig:central}
    \vspace{-10pt}
\end{figure}

In this section, we demonstrate the viability of MetaPoison against networks trained from scratch. For consistency, we focus on the same dog-bird and frog-plane class pairs as in previous work. To be thorough, however, we did a large study of all possible class pairs (appendix \S\ref{sec:allpairs}) and showed that these two class pairs are representative in terms of poisoning performance. We also found that even within the same poison-target class pair, different target images resulted in different poisoning success rates (appendix \S\ref{sec:targids}). Therefore, for each class pair, we craft 10 sets of poisons targeting the corresponding first 10 image IDs of the target class taken from the CIFAR-10 test set and aggregate their results. Finally, different training runs have different results due to training stochasticity (see appendix \S\ref{sec:trainingcurves} for training curves and \S\ref{sec:stability} for stability tests). Therefore, for each set of poisons, we train 6 victim networks with different random seeds and record the target image's label inferred by each model. In all, there are 60 labels, or votes: 6 for each of 10 different target images. We then tally up the votes for each class. For example, Figure \ref{fig:central} (lower left) shows the number of votes each label receives for the target birds out of 60 models. In unpoisoned models, the true class (bird) receives most of the votes. In models where just 1\% of the dataset is poisoned, the target birds get incorrectly voted as dog a majority of the time. Examples of some poison dogs along with their clean counterparts, as well as one of the target birds, are shown in Figure \ref{fig:central} (top left). More in appendix \S\ref{sec:morepoisons}. Note that a poison budget of 0.001\% is equivalent to zero poisons as the training set size is 50k.
In Figure \ref{fig:central} (top right), we repeat the experiments for multiple poison budgets and network architectures. Success rates of 40-90\% for a poison budget of 1\% are obtained for all architectures and class pairs we consider. ResNet20 achieves 72\% success with a 1\% budget on the dog-bird pair. The success rates drop most between 1\% and 0.1\%, but remain viable even down to 0.01\% budget. Remarkably, even a single perturbed dog can occasionally poison ResNet20. We also attempt using poisons crafted via feature collision (FC) for dog-bird. At 0\% success across all budgets, the failure of FC to work in train-from-scratch settings is elucidated. In Figure \ref{fig:central} (bottom right), we verify that our poisons cause negligible effect on overall model performance except at 10\% poison budget.
%
%

%

    
%

\vspace{-4pt}
\subsection{Robustness and transferability}
\label{sec:robustness}
\vspace{-4pt}

\begin{wrapfigure}[22]{r}{0.4\textwidth}
\vspace{-0pt}
\includegraphics[width=0.4\textwidth,trim=2cm 0cm 2cm 0cm, clip]{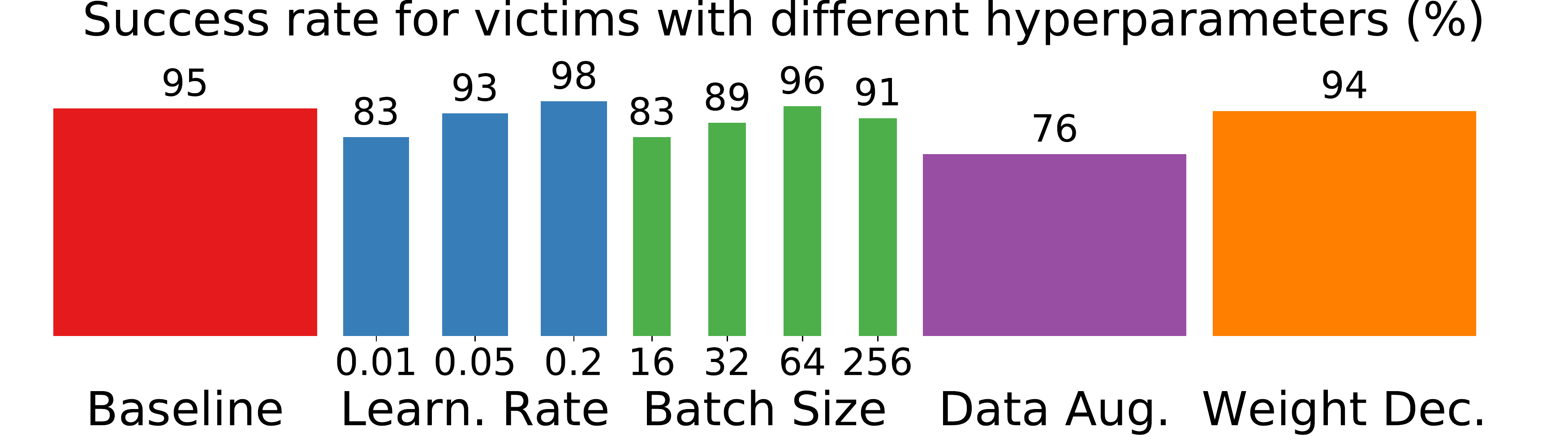}
\includegraphics[width=0.38\textwidth,trim=3cm .1cm .9cm .1cm, clip]{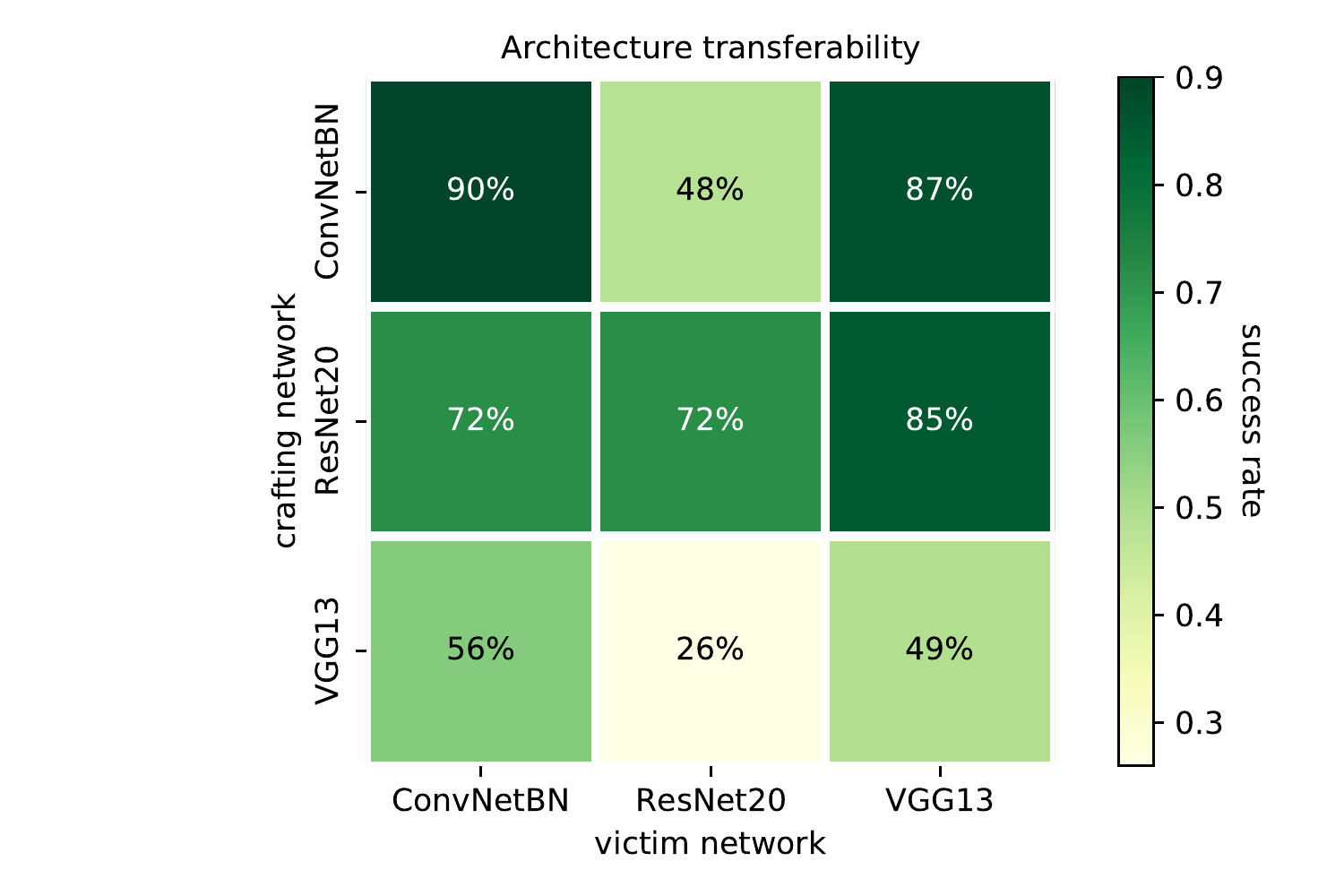}
\vspace{-11pt}
\caption{(Top) Success rate on a victim ConvNetBN with different training settings. (Bottom) Success rate of poisons crafted on one architecture and evaluated on another.}
\label{fig:robustness}
\end{wrapfigure}

So far our results have demonstrated that the crafted poisons transfer to new initializations and training runs. Yet often the exact training settings and architecture used by the victim are also different than the ones used to craft the poisons. We investigate the robustness of our poisons to changes in these choices. In Figure \ref{fig:robustness} (top), we train victim models with different training settings, like learning rate, batch size, and regularizers, on poisons crafted using ConvNetBN with a single baseline setting (0.1 learning rate, 125 batch size, no regularization). With a poison budget of 1\%, poison dogs were crafted for 10 different target birds and 30 victim models were trained per target. Our results show that the poisons are overall quite robust to changes.
Data augmentation (standard CIFAR-10 procedure of random crops and horizontal flips) and large changes in learning rate or batch size cause some, but not substantial degradation in success. The robustness to data augmentation is surprising; one could've conceived that the relatively large changes by data augmentation would nullify the poisoning effect.

Next we demonstrate architectural transferability in our trained-from-scratch models. In Figure \ref{fig:robustness} (bottom), using the same baseline experimental settings and poison budget as above, we craft poisons on one architecture and naively evaluate them on another. Despite ConvNetBN, VGG13, and ResNet20 being very different architectures, our poisons transfer between them. Interestingly the attack success rate is non-symmetric. Poisons created on VGG13 do not work nearly as well on ResNet20 as ResNet20 poisons on VGG13. One explanation for this is that VGG13 does not have batch normalization, which may have a regularizing effect on poison crafting. In practice, the attacker can choose to craft on the strongest architecture at their disposal, e.g. ResNet20 here, and enjoy high transferability, e.g. $>70\%$ here, to other architectures.

\begin{figure}[b]
    \vspace{-11pt}
    \centering
    \includegraphics[width=.697\textwidth]{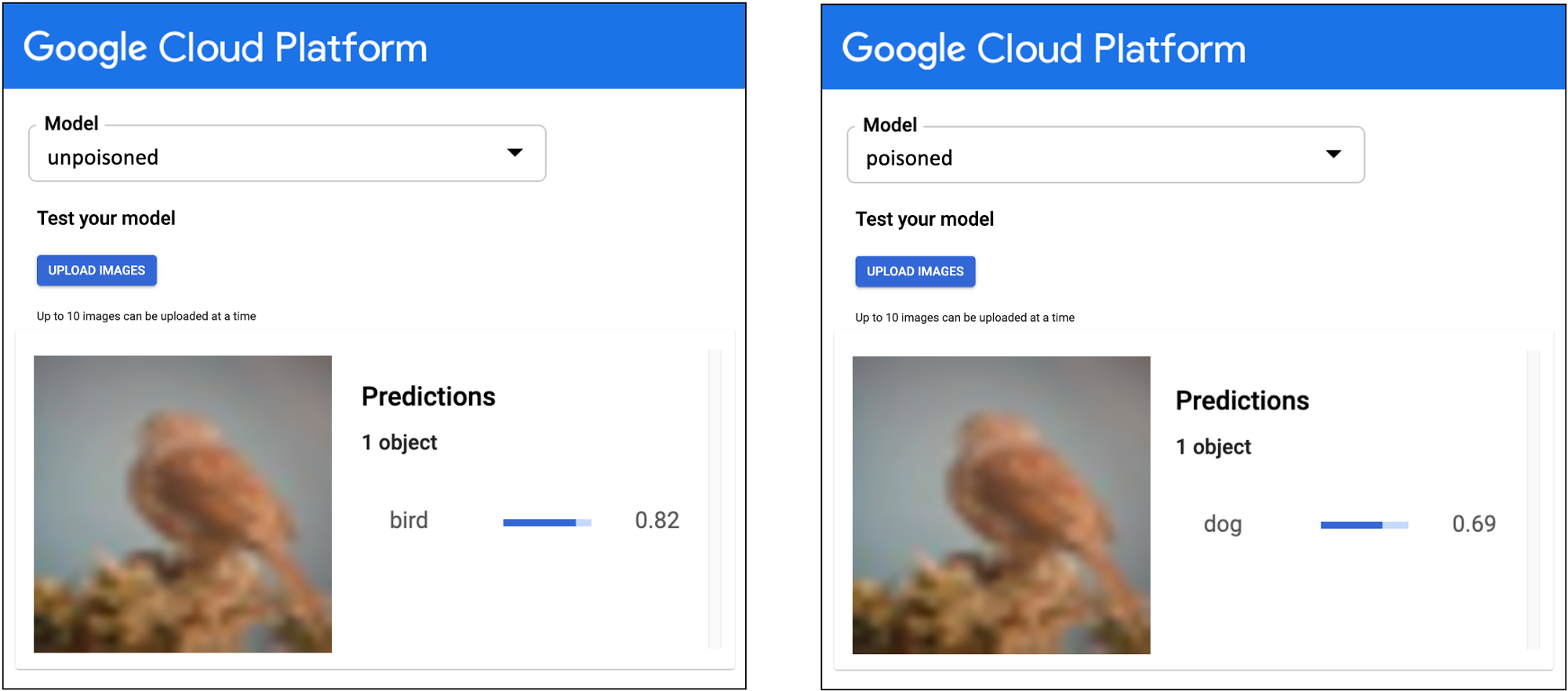}
    \quad
    \includegraphics[width=.258\textwidth,trim=0cm .22cm 0cm .09cm, clip]{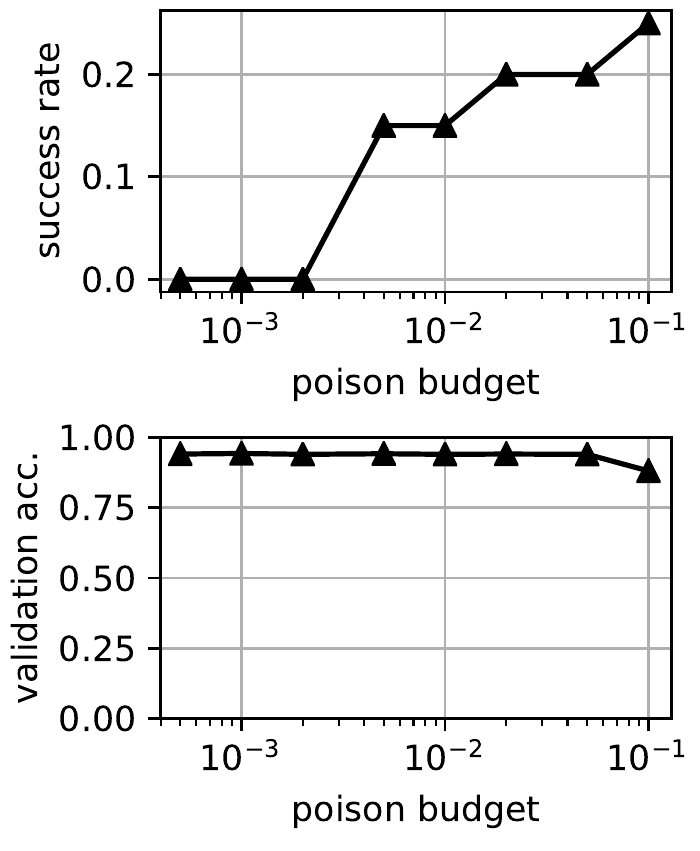}
    \caption{Data poisoning Google Cloud AutoML Vision models. Web UI screenshots of prediction results on target bird by Cloud AutoML models trained on (Left) clean and (Middle) poisoned CIFAR-10 datasets. Portions of the screenshot were cropped and resized for a clearer view. (Right) Success rates and validation accuracies averaged across 20 targets and training runs.} 
    \label{fig:automl}
    \vspace{-10pt}
\end{figure}

\vspace{-4pt}
\subsection{Poisoning Google Cloud AutoML API}
\label{sec:automl}
\vspace{-4pt}

We further evaluate the robustness of MetaPoison on the Google Cloud AutoML API at \url{cloud.google.com/automl}, a real-world, enterprise-grade, truly \textit{black-box} victim learning system. Designed for the end-user, Cloud AutoML hides all training and architecture information, leaving the user only the ability to upload a dataset and specify wallclock training budget and model latency grade. For each model, we upload CIFAR-10, poisoned with the same poison dogs crafted earlier on ResNet20, and train for 1000 millinode-hours on the mobile-high-accuracy-1 grade. After training, we deploy the poisoned model on Google Cloud and upload the target bird for prediction. Figure \ref{fig:automl} shows web UI screenshots of the prediction results on unpoisoned (left) and poisoned (middle) Cloud AutoML models. MetaPoison works even in a realistic setting such as this. To quantify performance, we train 20 Cloud AutoML victim models, 2 for each of the first 10 target birds, and average their results in Figure \ref{fig:automl} (right) at various budgets. At poison budgets as low as 0.5\%, we achieve success rates of $>$15\%, with little impact on validation accuracy. These results show that data poisoning presents a credible threat to real-world systems; even popular ML-as-a-service systems are not immune to such threats. 

\vspace{-20pt}
\subsection{Versatility to alternative poisoning schemes}
\vspace{-4pt}

Thus far we have discussed targeted poisoning attacks under a \textit{collision} scheme: inject poisons from class $y_{p}$ to cause a specific instance in class $y_{t}$ to be classified as $y_{p}$. In other words, the adversarial class is set to be the poison class, $y_\textrm{adv} = y_{p}$. This was the only scheme possible under the feature collision method. It is however only a subset of the space of possible schemes $\mathcal{Y}_{scheme}: (y_p, y_t, y_\textrm{adv})$. Since MetaPoison learns to craft poison examples directly given an outer objective $\mathcal{L}(x_t, y_\textrm{adv}; \theta^*(X_p))$, it can accomplish a a wide range of alternative poisoning objectives.
In addition to showing MetaPoison's success on existing, alternative schemes such as \textit{multi-target} or \textit{indiscriminate} poisoning \citep{steinhardt2017poison, munoz-gonzalez_towards_2017} in appendix \S\ref{sec:multitarget}, here we demonstrate MetaPoison's versatility on two alternative, never-before-demonstrated schemes.

\begin{figure}[t!]
    \vspace{-25pt}
    \centering
    \includegraphics[width=.285\textwidth,valign=t,trim=0 0 0 .05cm,clip]{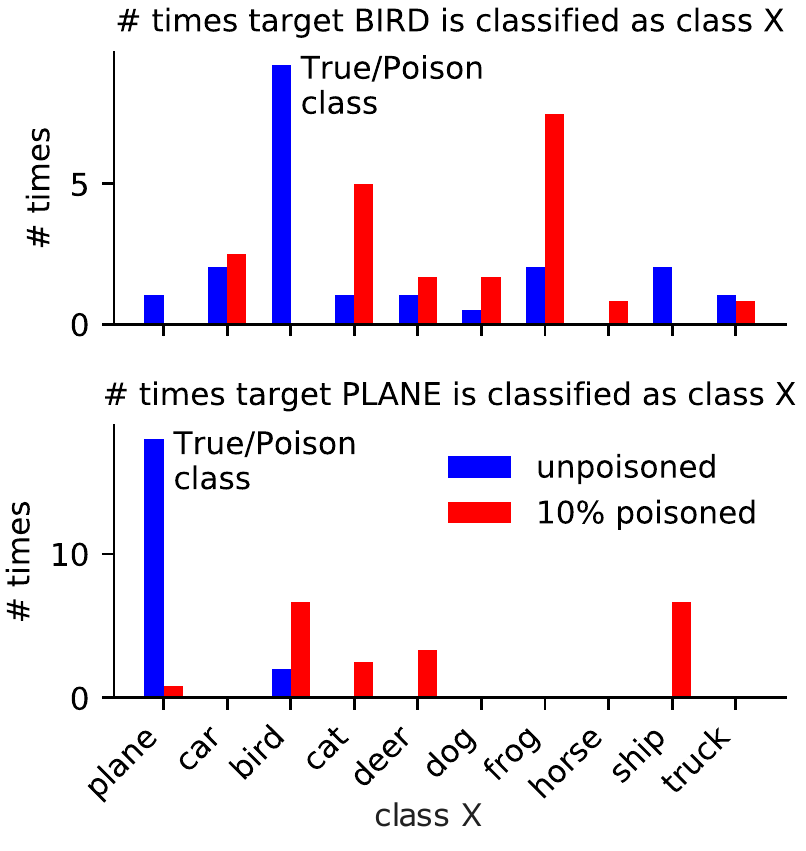}
    \quad\quad
    \includegraphics[width=.365\textwidth,valign=t,trim=0 0 0 .2cm,clip]{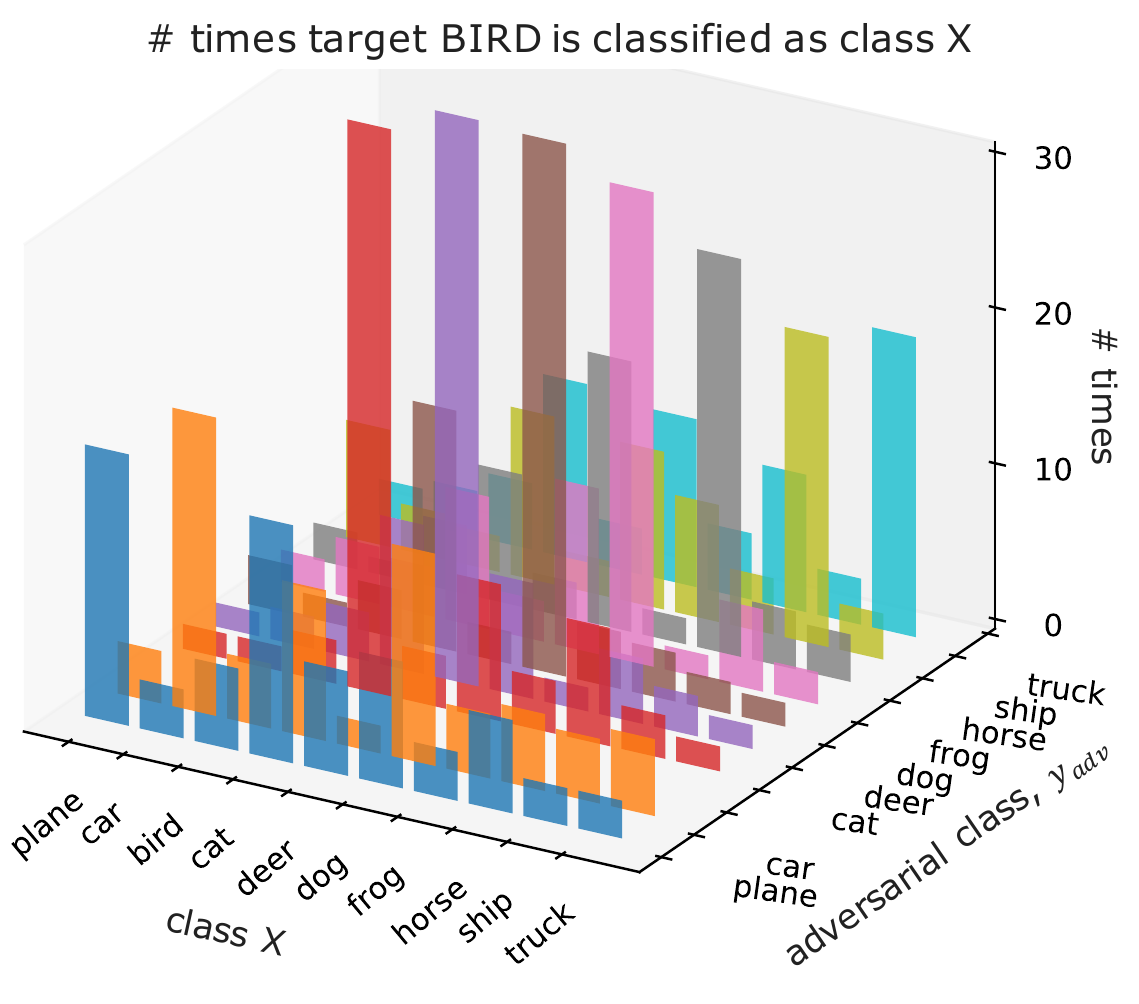} 
    \vspace{-7pt}
    \caption{Alternative poisoning schemes. (Left) Self-concealment: images from the same class as the target are poisoned to ``push" the target out of its own class. (Right) Multiclass-poisoning: images from multiple classes are poisoned to cause the target bird to be classified as a chosen adversarial class $y_\textrm{adv}$.}
    \label{fig:schemes}
    \vspace{-8pt}
\end{figure}

\textbf{Self-concealment scheme}: Poisons are injected to cause a target image in the \textit{same} class to be misclassified, i.e. $y_p = y_t \neq y_\textrm{adv}$. E.g. attackers submit altered photos of themselves to a face identification system to \textit{evade} being identified later. To implement this, we simply change the adversarial loss function to $\mathcal{L}_\textrm{adv}(X_p) = -\log\left[1 - p_{\theta^*(X_p)}(x_t, y_t)\right]$ so that higher misclassification of the target lowers the loss. We evaluate the self-concealment scheme on two poison-target pairs, bird-bird and airplane-airplane. We use a poison budget of 10\% and like in Figure \ref{fig:central} (bottom left), tally the labels given to the target bird or plane by 20 victim models. Figure \ref{fig:schemes} (left) shows histograms of these tallies. For unpoisoned models, the true label receives the clear majority as expected, while for poisoned models, the votes are distributed across multiple classes without a clear majority. Usually, the true label (bird or airplane) receives almost no votes by poisoned models. Using definition of success as misclassification of the target, the success rates are 100\% and 95\% for bird and airplane, respectively.

\textbf{Multiclass-poison scheme}: In cases where the number of classes is high, it can be difficult to assume a large poison budget for any single class. E.g. if there are 1000 classes (balanced), the max poison budget for poisoning only a \textit{single} class is 0.1\%, which may not be always enough poisons.
One solution is to craft poisons in \textit{multiple} classes that act toward the same goal.
Here, we craft poisons uniformly distributed across the 10 CIFAR-10 classes with a 10\% total budget, or 1\% poison budget in each class. Our goal is to cause a target bird to be assigned a chosen incorrect, adversarial label $y_\textrm{adv}$. Like before, we tally the predictions over 60 victim models. Figure \ref{fig:schemes} (right) shows 9 histograms.
Each histogram shows the distribution of the 60 predictions of victim models poisoned with a particular choice of adversarial label $y_\textrm{adv}$.
For example, the blue, frontmost histogram in Figure \ref{fig:schemes} shows how the target bird image is perceived by 60 victim models that are poisoned with an adversarial class of plane---the bird is (mis)perceived as a plane most of the time. In general, for most of the 9 histograms, the class that receives the most votes is the adversarial class. On average, the adversarial class claims 40-50\% of the votes cast, i.e. 40-50\% success. This attack shows that it's possible to use poisons from multiple classes to arbitrarily control victim label assignment.

%% file: conclusion.tex
\vspace{-8pt}
\section{Conclusion}
\vspace{-5pt}

MetaPoison finds dataset perturbations that control victim model behavior on specific targets. It out-performs previous clean-label poisoning methods on fine-tuned models, and achieves considerable success---for the first time---on models trained from scratch. It also enables novel attack schemes like self-concealment and multiclass-poison. Unlike previous approaches, the poisons are \textit{practical}, working even on industrial black-box ML-as-a-service models. We hope MetaPoison establishes a baseline for data poisoning work and promotes awareness of this very real and emerging threat vector.


%% file: impact.tex
\section{Broader Implications}

Data lies at the heart of modern machine learning systems. The ability of MetaPoison to attack real-world systems is should raise awareness of its broader implications on computer security and data/model governance. While a full discussion should involve all stakeholders, we provide here some initial comments. First, data and model governance is of utmost importance when it comes to, among other things, mitigating data poisoning. \citet{elie2018attacks} provides some common-sense steps to take when curating a training set. For example, one should ensure that no single source of data accounts for a large fraction of the training set or even of a single class, so as to keep the poison budget low for a malicious data contributor.
Second, it is easier to defend against wholesale model skewing attacks which aim to reduce overall model performance or to bias it toward some direction. Targeted attacks such as ours, on the other hand, are far more difficult to mitigate, since the overall model behavior is unchanged and the target input on which the model's behavior \textit{is} changed is not known to the victim.
Systems should rely on additional auxiliary measures, such as interpretability techniques \citep{kim2017interpretability}, to make security-critical decisions.
Third, at the moment, the computational power required to craft MetaPoison examples exceeds that of evasion attacks by a large margin. This provides researchers time to design mitigation strategies before it becomes a dominant threat to real-world systems, as well as study robust learning techniques that leverage, e.g., computational hardness \citep{mahloujifar2019can}.
As a final note, data poisoning techniques are not limited to nefarious uses. For example, it can be used for copyright enforcement as discussed in \S\ref{sec:intro} and similar to the concept of ``radioactive data'' \citep{sablayrolles2020radioactive}. Another non-nefarious use case is privacy protection \citep{shan2020fawkes}.

%% file: acknowledgments.tex
\section*{Acknowledgments}
\url{www.comet.ml} supplied necessary tools for monitoring and logging of our large number of experiments and datasets and graciously provided storage and increased bandwidth for the unique requirements of this research project. The authors had neither affiliation nor correspondence with the Google Cloud AutoML Vision team at the time of obtaining these results. Goldstein and his students were supported by the DARPA's GARD program, the DARPA QED for RML program, the Office of Naval Research MURI Program, the AFOSR MURI program, and the Sloan Foundation. LF was supported in part by LTS through Maryland Procurement Office and by the NSF DMS 1738003 grant. Taylor was supported by the Office of Naval Research. 
%

%% file: supplementary.tex
\clearpage

%



\vbox{%
    \hsize\textwidth
    \linewidth\hsize
    \vskip 0.1in
  \hrule height 4pt
  \vskip 0.25in
  \vskip -\parskip%
    \centering
    {\LARGE\bf Appendix\par}
  \vskip 0.29in
  \vskip -\parskip
  \hrule height 1pt
  \vskip 0.09in%
}

\appendix

\section{Poison crafting curves}
\label{sec:craftsteps}

Our poisons in the main paper were all crafted with 60 outer steps, also called craft steps. Here we investigate the outer optimization process in more depth and show the potential benefits of optimizing longer. As a testbed, we consider poison frogs attacking a target airplane with a poison budget of 10\%. During the crafting stage, the adversarial loss--we use the \citet{carlini2017towards} loss here--is the objective to be minimized. This loss has the property that when it is less than zero, the target is successfully misclassified as the adversarial class. Conversely, when it is greater than zero, the target is classified into a class other than the adversarial class.

\begin{figure}[h!]
    \centering
    \includegraphics[width=0.4\textwidth]{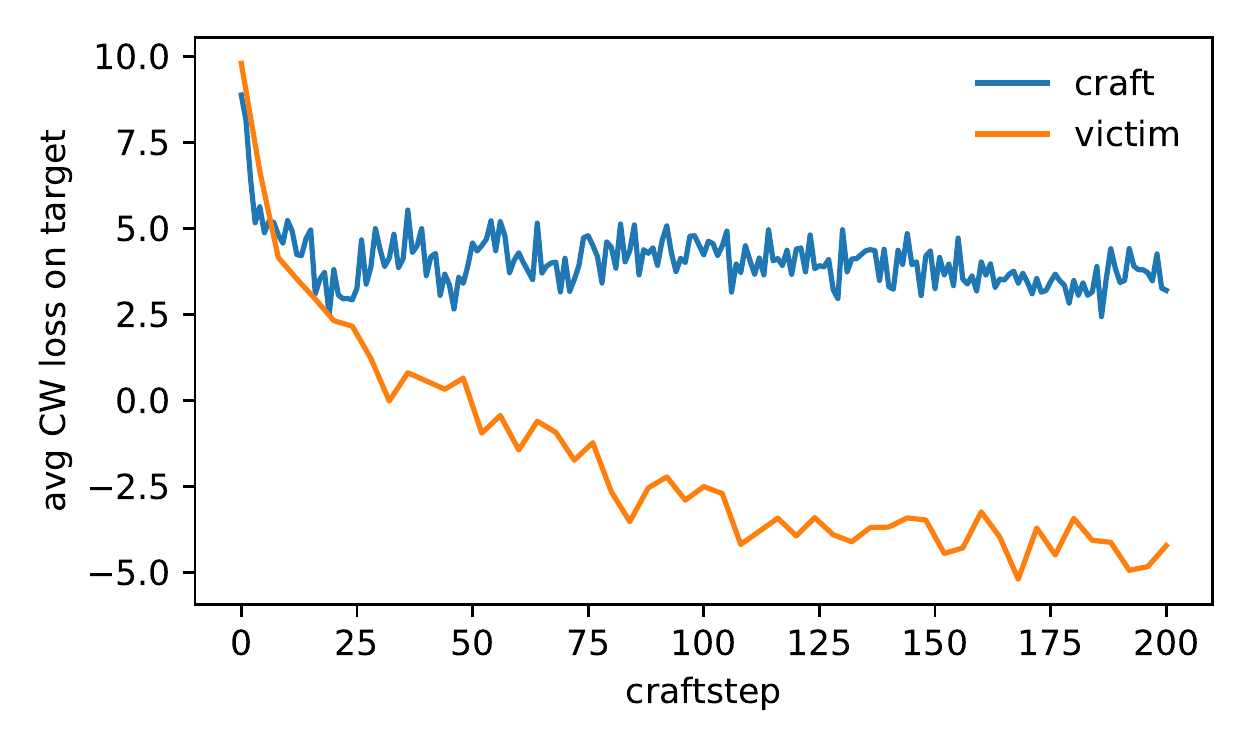} \\
    \vspace{-1pt}
    \includegraphics[width=0.4\textwidth]{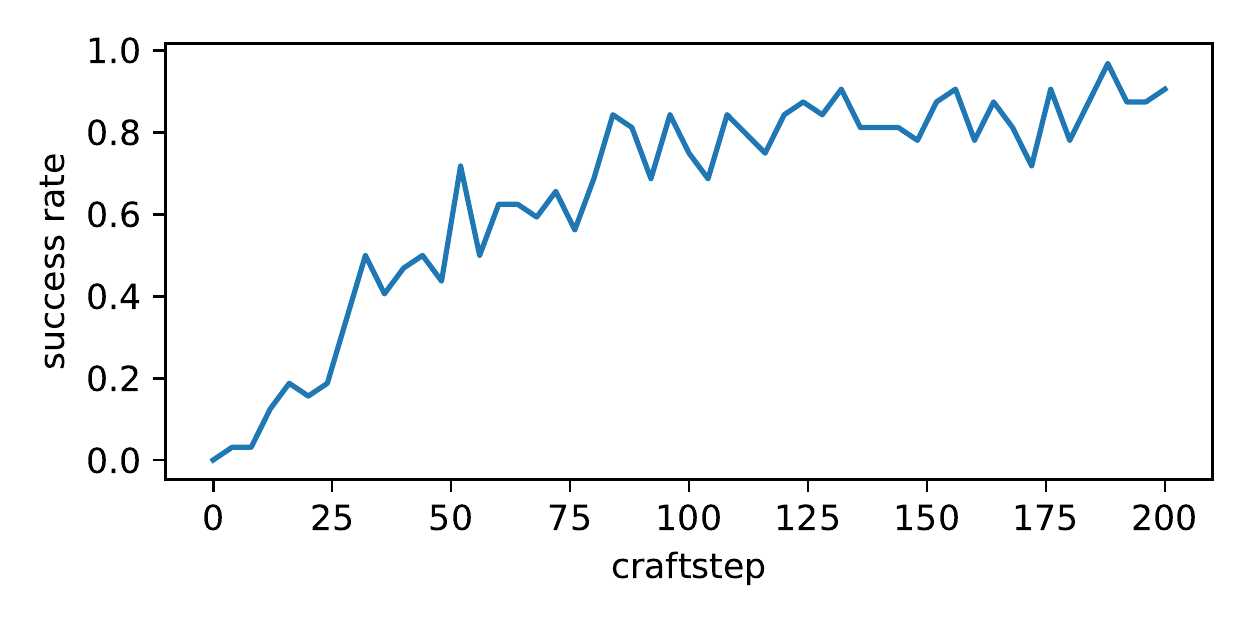}
    \vspace{-3pt}
    \caption{Ablation study on the number of craftsteps. (Top) The crafting adversarial loss (blue line), which is averaged across all 24 models in the ensemble, is the objective to be minimized in the outer loop of the bi-level optimization problem.
    We save the state of the poisons at every several craftsteps, fully train 20 victim models from scratch on each of those poisons, and plot the average adversarial loss on the target across those victim models (orange line). (Bottom) Attack success rate across the 20 victim models for each craft step.}
    \label{fig:craftstep}
\end{figure}

The blue line in Figure \ref{fig:craftstep} (top) shows the adversarial loss averaged over all the surrogate models during the crafting stage. It rapidly decreases up to craftstep 25 and then plateaus. It never sinks below zero, which means that inserting these poisons into a minibatch will not cause the model to misclassify the target two look-ahead SGD steps later, on average. However, it belies the fact that the cumulative effect of the poisons will collectively influence the model to misclassify the target after many SGD steps. Indeed, the fact that the adversarial loss (blue line) is decreased after 25 craft steps from $\sim$9 to $\sim$4 is an indication that the poisons provide a small nudge to the model toward misclassifying the target even after two look-ahead SGD steps, as compared to having no poisons.

The orange line in Figure \ref{fig:craftstep} (top) shows the adversarial loss on the target image on poisoned victim models at each stage of crafting. To obtain this curve, we saved the state of the poisons every several craft steps, and trained 20 victim models from scratch on each of them. Interestingly, even though the crafting adversarial loss (blue line) plateaus, the effectiveness of the poisons continues to increase with the number of craft steps even up to 200 steps. Therefore, one cannot judge from the crafting curve alone how well the poisons will do during victim evaluation. Finally, Figure \ref{fig:craftstep} (bottom) shows the corresponding attack success rate for the poisons at each craft step.

\section{Victim training curves}
\label{sec:trainingcurves}

In the main paper, we reported the attack success rates and validation accuracy at the \textit{end} of victim training. In this section, we take a closer look at the effect of data poisoning at each step of training.

\begin{figure}[h!]
    \centering
    \includegraphics[width=0.43\textwidth]{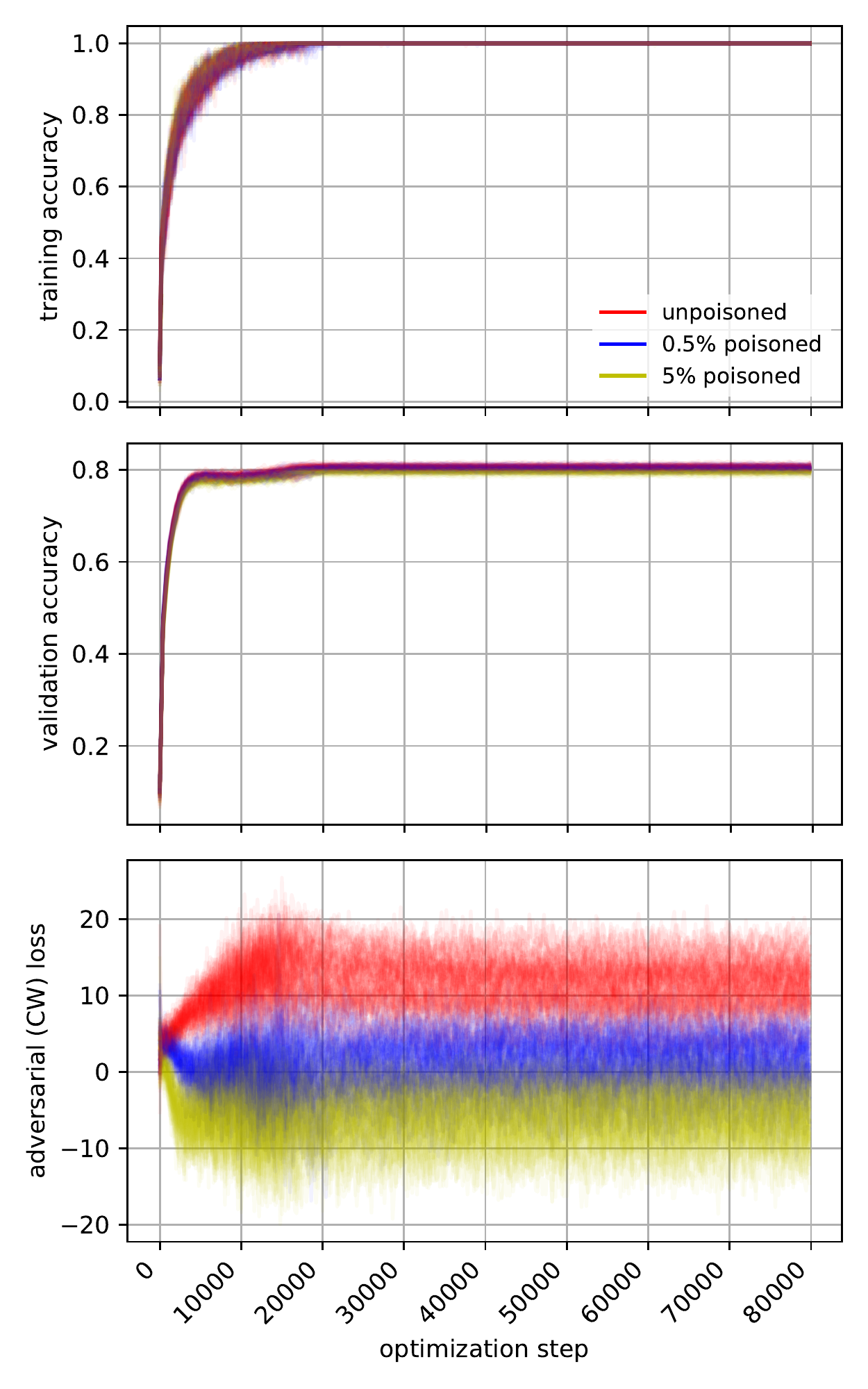}
    \caption{Training curves from scratch with different random seeds on poisoned and unpoisoned datasets over 200 epochs on ResNet20. (Top) Accuracy on training set perfectly overfits to CIFAR-10 after about 20k optimization steps, or 50 epochs. (Middle) Validation accuracy curve looks the same regardless of whether the dataset is poisoned or not. (Bottom) Carlini-Wagner (CW) adversarial loss on specific target bird (ID 5) as a function of optimization step. CW loss above zero indicates the target bird is classified correctly, while below zero indicates the target bird is misclassified as a dog. Unpoisoned models have adversarial loss entirely above zero, while 5\% poisoned models have adversarial loss entirely below zero. 0.5\% poisoned models have CW loss straddling both sides of zero.}
    \label{fig:victimcurves}
\end{figure}

We again use the dog-bird class pair as our prototypical example and we randomly select target bird with ID 5. We train ResNet20 models with 3 different poisoning levels: unpoisoned, poisoned with 0.5\% budget, and poisoned with 5\% budget. Since the training of each victim model is inherently stochastic and we desire to see the overall effect of poisoning, we train 72 victim models with different seeds for each of these 3 poisoning levels. Figure \ref{fig:victimcurves} displays all 72 curves for each poisoning level. The training accuracy curves, in Figure \ref{fig:victimcurves} (top), show the models quickly overfitting to the CIFAR10 dataset after about 20k optimization steps, or 50 epochs. The rate of convergence is equal for all 3 poisoning levels. Likewise, the validation accuracy curves, Figure \ref{fig:victimcurves} (middle), converge to about 80\% after 20k steps and are also indistinguishable between poisoning levels. These curves show that it is impossible to detect the presence of poisoning through looking at training or validation curves.

Next, we look at the evolution of the adversarial loss, or \citet{carlini2017towards} loss, over optimization step in Figure \ref{fig:victimcurves} (bottom). Recall that in the \citet{carlini2017towards} loss, negative values correspond to attack success while positive values correspond to attack failure. Note also that, under almost all practical scenarios, the victim does not see this curve since they are unaware of the target image chosen by the adversary.

At the start, epoch 0, the adversarial loss of all models are at roughly the same level. As training proceeds, the adversarial loss trifurcates into 3 distinct levels corresponding to the 3 poisoning levels. The unpoisoned models see increasing adversarial loss up to fully positive values (perfect attack failure) of around 12 before they plateau, while the high 5\% poisoned models see decreasing adversarial loss down to mostly negative values (near-perfect attack success) of around $-6$ before plateauing. The moderate 0.5\% poisoned models see slight decrease in adversarial loss and hover around zero (some attack success) for the remainder of training. Compared to the training and validation curves, these adversarial loss curves fluctuate a lot both between optimization steps as well as between models. This is expected since they are the loss of a single image rather than an aggregate of images. Despite the fluctuation, however, the effect of different poisoning levels on the attack outcome is very clear.

\section{Performance on other poison-target class pairs}
\label{sec:allpairs}
\begin{figure}
    \centering
    \includegraphics[width=0.49\textwidth]{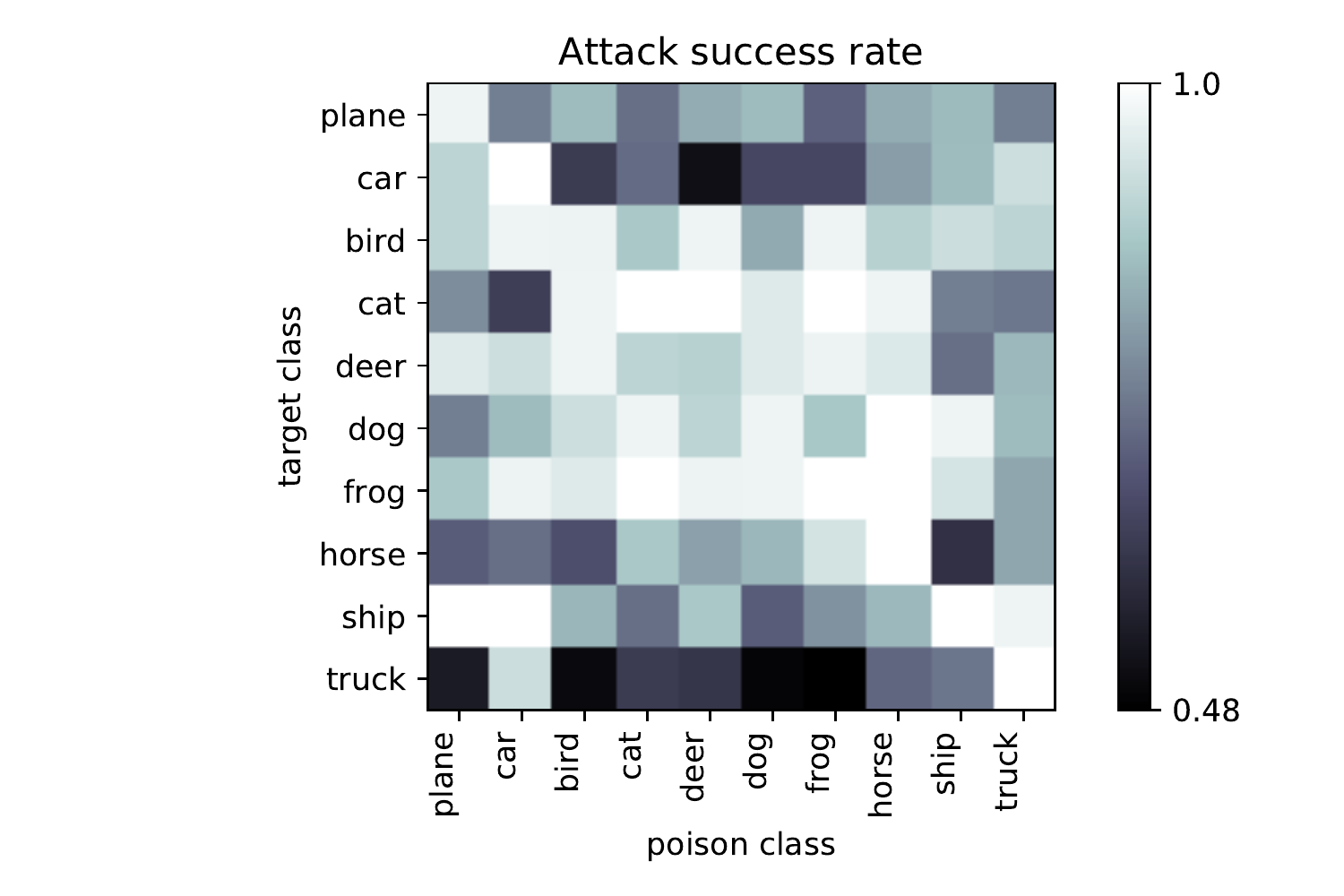}
    \caption{Success rates for all possible poison-target class pairs. Each success rate is the average of the first 5 unique targets with 2 victim training runs per unique target.}
    \label{fig:allpairs}
    \vspace{-5pt}
\end{figure}

In the main paper, we primarily mimicked the two exemplary poison-target class pairs (dog-bird, frog-airplane) from previous work in \citet{shafahi2018poison}. To ensure that our results do not just happen to work well on these two pairs but rather works well for all class pairs, we perform a large study on all 100 possible poison-target pairs in CIFAR-10, shown in Figure \ref{fig:allpairs}.

For each pair, we craft with a poison budget of 10\%, target the first 5 target IDs for that particular target class, and run 2 victim trainings from scratch for each pair, allowing the reported success rate to result from the average of 10 victim models. To enable such a large study within our computational runtime constraints, we use only 10\% of the CIFAR-10 dataset as our training set. This is justified since we are interested here in the relative performance of different class pairs with respect to our exemplary class pairs (dog-bird, frog-airplane) on which we did full CIFAR-10 studies in the main paper.

The results show that poisoning can be successful under \textit{all} class pair situations. Our exemplary pairs, dog-bird and frog-airplane, have average poisoning vulnerability relative to all the other pairs, with the dog-bird slightly more vulnerable than frog-airplane. The most difficult target class on which to cause misclassification is truck, while the most easy is frog. The least powerful poison class is truck, while the most powerful is tied between car, cat, deer, frog, and horse. The high success rates along the diagonal trivially indicate that it is easy to cause the target to be classified into the correct class.

\section{Differences in success rates amongst different targets}
\label{sec:targids}

\begin{figure}
    \vspace{-10pt}
    \centering
    \includegraphics[width=0.43\textwidth]{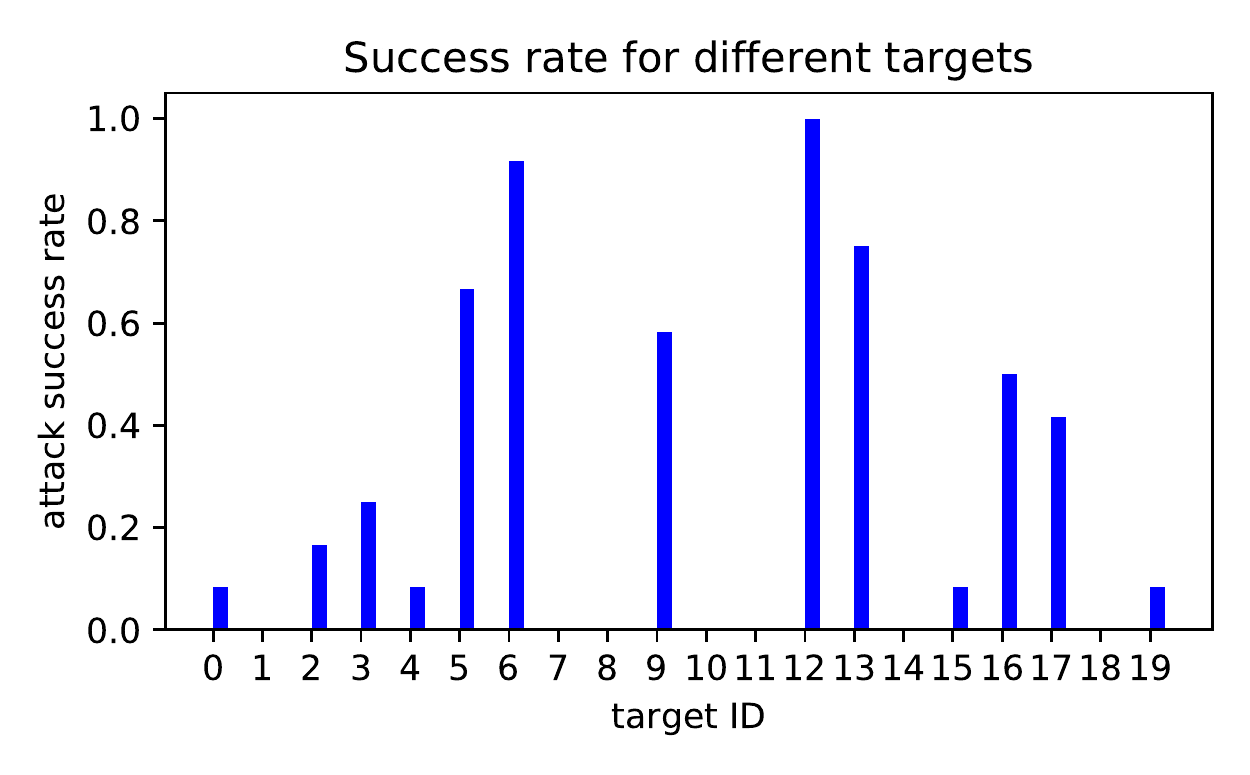}
    \vspace{-10pt}
    \caption{Success rates for the first 20 unique target airplanes for a poison frog target airplane situation. Each success rate is the average of 12 victim training runs.}
    \label{fig:targetids}
\end{figure}

It is also informative to see how the success rate varies amongst different choices of the target image for a fixed target class. Even though the target class is the same, different images within that class may have very different features, making it harder or easier for the poisons to compromise them. In Figure \ref{fig:targetids}, we plot the attack success rates for the first 20 unique target airplanes when attacked by poison frogs. Each success rate is the result of 20 victim training runs. Indeed, the success rate is highly variable amongst different target images, indicating that the poisoning success is more dependent on the \textit{specific} target image that the adversary wishes to attack rather than the choice of poison-target class pair.

\section{Ablation study on ensemble size}
\label{sec:ablationensembling}
\begin{figure}
    \vspace{-20pt}
    \centering
    \includegraphics[width=0.49\textwidth]{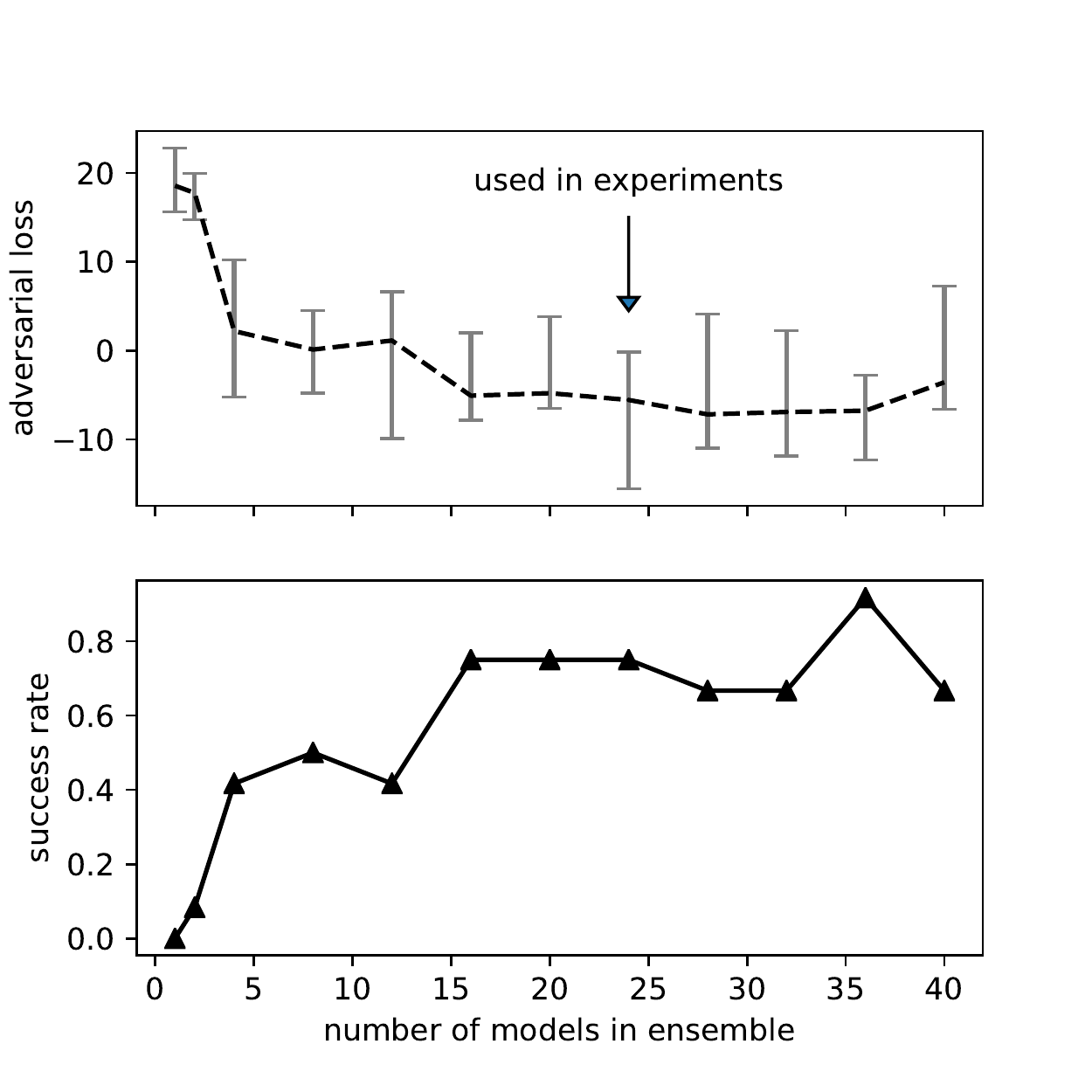}
    \vspace{-10pt}
    \caption{Ablation study on surrogate ensemble size.}
    \label{fig:ensemble}
    \vspace{-0pt}
\end{figure}

Throughout the main paper, we have used an ensemble size of 24 surrogate models, reasoning that ensembling of models at different epochs encourages the poisons to be effective for all network initializations and training stages. Here, we perform an ablation study of poisoning success against ensemble size in Figure \ref{fig:ensemble}. Poisons crafted without ensembling (ensemble size of 1) are ineffective, while success rate trends upward as ensemble size increases. We also show empirically that our ensemble size of 24 lies where success rate saturates, balancing poison success w/ computational efficiency.

\section{Ablation study on reinitialization}
\label{sec:ablationreinit}
We substantiate the claim in the paper (\S2.2) that network reinitialization of the surrogate models contributes to making more effective poisons by running an ablation study where instead of reinitializing the surrogate networks every sentinel number of epochs, we keep their original initialization fixed throughout the crafting process. Over 100 victim training runs, the average success rate of poisons crafted via fixed initialization was 51\% while the baseline of reinitialization achieved 60\%, showing that reinitialization causes a modest but significant enhancement poisoning efficacy. Poisons crafted on fixed initialization networks are less effective than their reinitialization counterparts.

\section{Indiscriminate and multi-targeted attacks}
\label{sec:multitarget}
\begin{figure}
    \centering
    \includegraphics[width=0.4\textwidth]{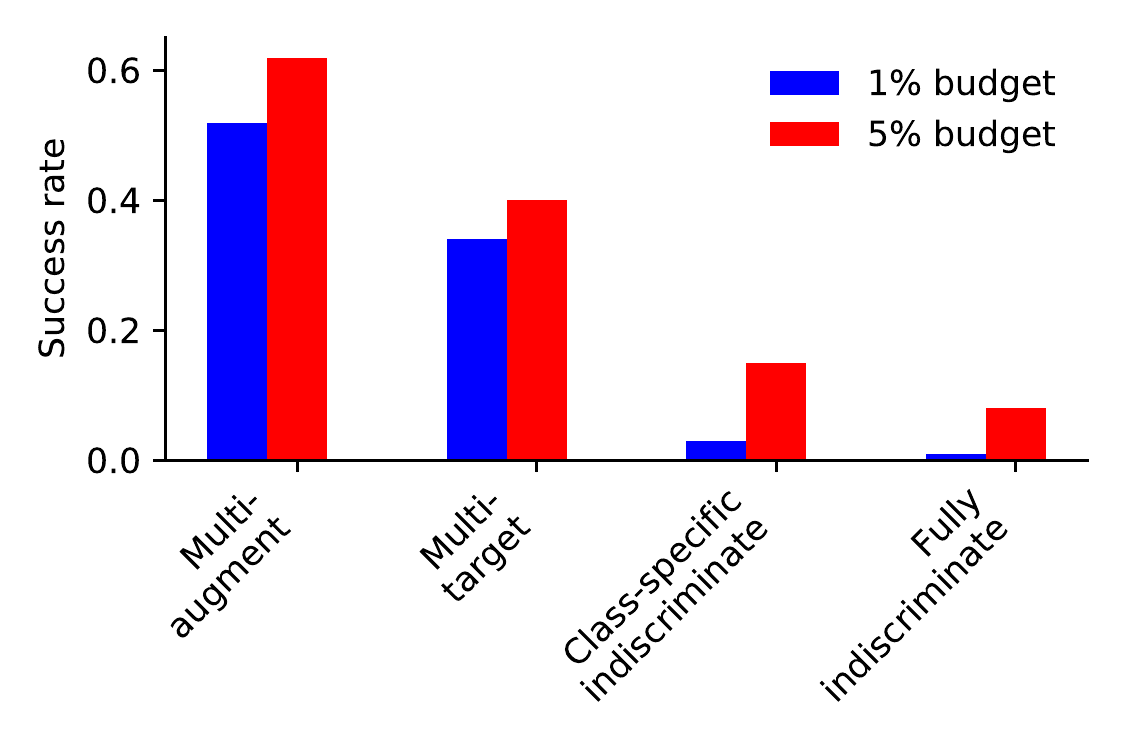}
    \vspace{-10pt}
    \caption{Results on various attack variants that involve more targets. Variants lie along the spectrum which goes from causing a single target to be misclassified (as studied in the main paper) to multiple targets to misclassifying indiscriminately.}
    \label{fig:indiscriminate}
    \vspace{-0pt}
\end{figure}

The paper focused poison attacks where the goal is to cause a single target instance to be misclassified since it is a straightforward and realistic scenario. However there are situations where the attacker may want to cause multiple targets to be misclassified, or take down the system by causing it to misclassify indiscriminately. Here we evaluate MetaPoison's effectiveness on four attack variants along the single-target spectrum/multi-target/indiscriminate spectrum, including 1. multiple (>10) augmentations of the same target object, 2. multiple (5) distinct target objects, 3. indiscriminate for a specific class (bird), i.e. error-specific, and 4. fully indiscriminate, i.e. error-generic. To adapt MetaPoison to the indiscriminate attack variants, we redefine the adversarial loss to be the average Carlini-Wagner loss over a random minibatch of target images sampled from a hold-out set, while for multi-target attack variants we redefine it to be the average loss over the multiple target images or over random augmentations. For indiscriminate attacks, we define success rate to be the amount of error increase caused by the poisons over the baseline unpoisoned error rate of the model. Figure \ref{fig:indiscriminate} shows the results for two poison budgets. MetaPoison performs modestly on the more indiscriminate attacks, a few percentage points of error increase. This could be attributed to the fact that the adversarial loss for indiscriminate attacks, which takes into account all images in a hold-out set, places quite a few constraints on the poisons. Meanwhile Figure \ref{fig:indiscriminate} shows it's possible to achieve double-digit success rates on the more targeted attacks since the constraint involves fewer images.

\section{Ablation study on perturbation magnitude}
\label{sec:pertmag}

\begin{figure}
    \centering
    \includegraphics[width=0.60\textwidth]{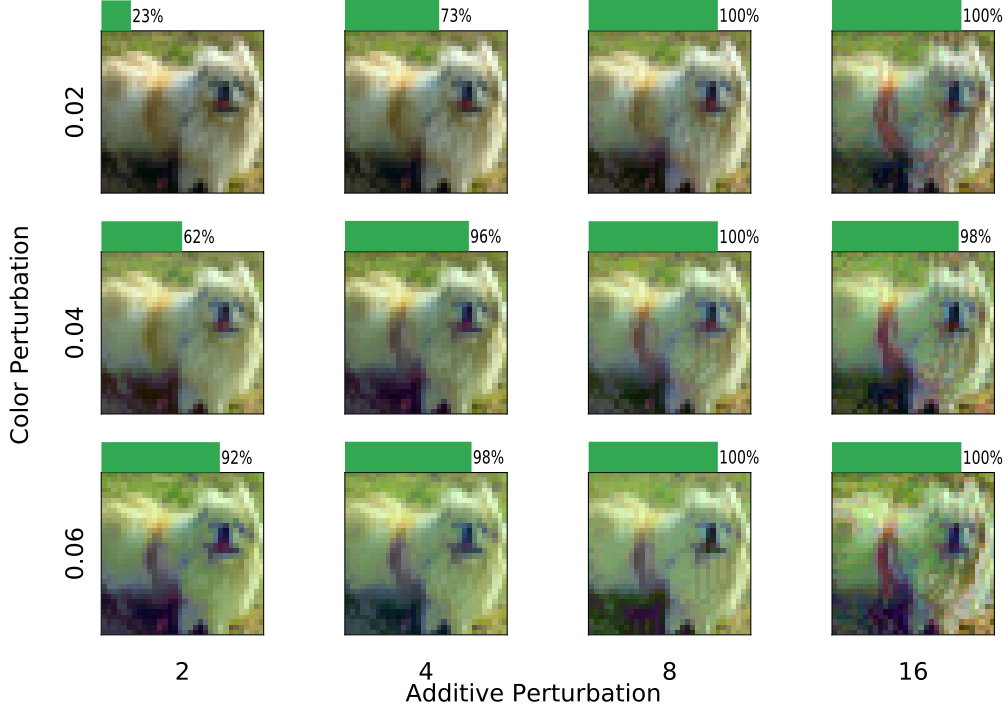}
    \vspace{-0pt}
    \caption{Ablation study of perturbation. We vary the strength of the attack by modifying the allowed $\ell_\infty$-perturbation $\epsilon$ (y-axis) and the color perturbation $\epsilon_c$ (x-axis) and show an exemplary poison image (from the batch of 1\% poison images). The green bars show attack success. Note that the baseline used in all other experiments in this paper is a color perturbation $\epsilon_c$ of $0.04$ and additive perturbation $\epsilon$ of $8$.} 
    \label{fig:ablation}
    \vspace{-10pt}
\end{figure}

We present an ablation study for different additive and color perturbation bounds in Figure \ref{fig:ablation} for one particular dog-bird attack (bird ID 0) with 1\% poison budget. While our experiments use modest values of $(\eps, \eps_c) = (8, 0.04)$, there is room to increase the bounds to achieve higher success without significant perceptual change as shown by an example poison dog in the figure. In contrast, even extremely minimal perturbations $(\eps, \eps_c) = (2, 0.02)$ can achieve notable poisoning.

\section{Ablation study on number of unroll steps used during crafting}
\label{sec:unrollsteps}

\begin{figure}
    \centering
    \vspace{5pt}
    \includegraphics[width=0.37\textwidth]{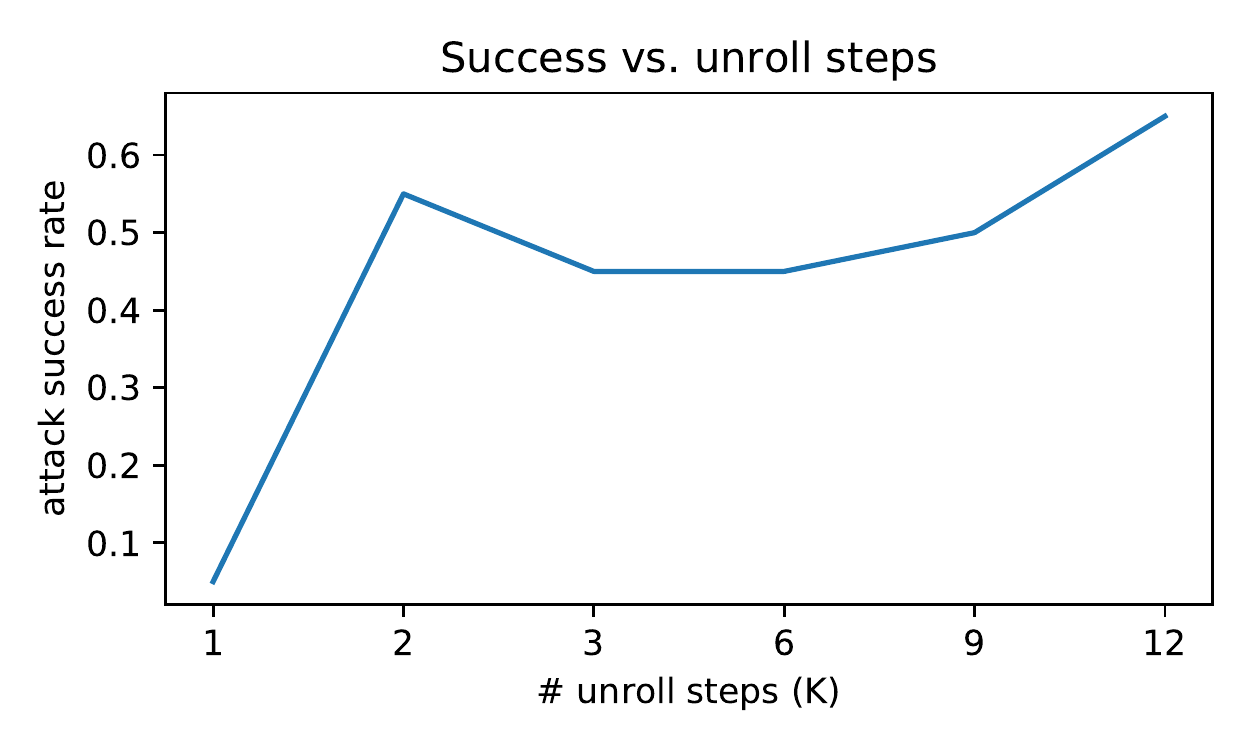}
    \caption{Ablation study on the number of unroll steps. Using a single unroll step during crafting will produce inferior poisons, but using a modest number between 2 and 9 seems to result in the same performance more or less. Even large numbers of unroll steps may improve the performance slightly.}
    \label{fig:unrolls}
\end{figure}

We now investigate how far we should look ahead during the crafting process, or for how many SGD steps we should unroll the inner optimization. It turns out a low-order approximation, or small number of unrolls, is sufficient when our ensembling and network reinitialization strategy is used. Figure \ref{fig:unrolls} shows the attack success rate for various choices of the number of unroll steps, or $K$ as defined in Algorithm 1. A single unroll step is insufficient to achieve high success rates, but having the number of unroll steps be between 2 and 9 seems to not affect the result much. At even higher number of unroll steps (12), the success rate increases slightly. We thus recommend using 2 unroll steps as it performs well while minimizing computational costs.

\section{Stability of poison crafting}
\label{sec:stability}
A reliable poison crafting algorithm should produce poisons of the same effectiveness under the same conditions with different random seeds. In nonconvex optimization, this is not always the case. MetaPoison's optimization procedure to craft poisons is certainly nonconvex, and it's unclear how the adversarial loss landscape looks like; therefore, in this section, we take a look at the stability of the poison crafting process.

\begin{figure}
    \centering
    \includegraphics[width=.4\textwidth]{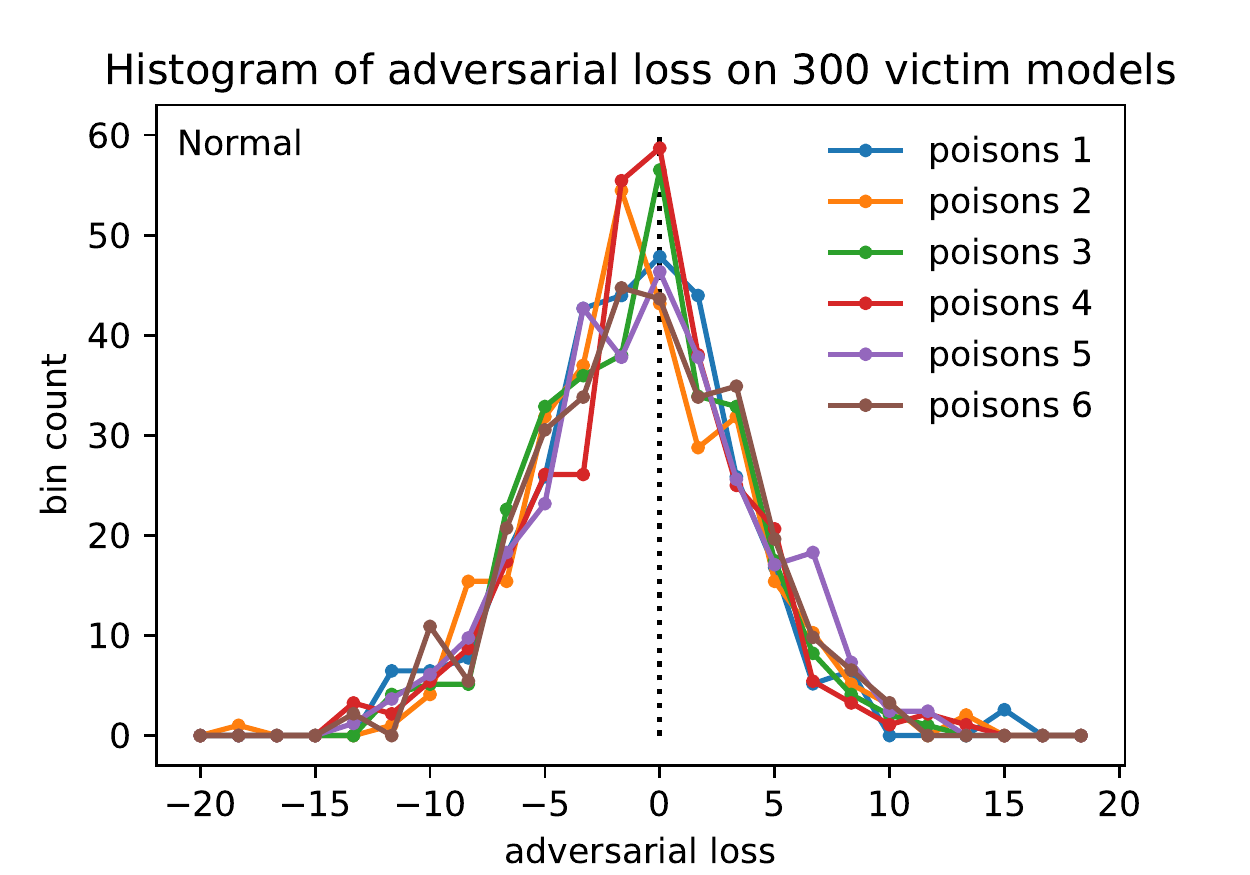} \\
    \includegraphics[width=.4\textwidth]{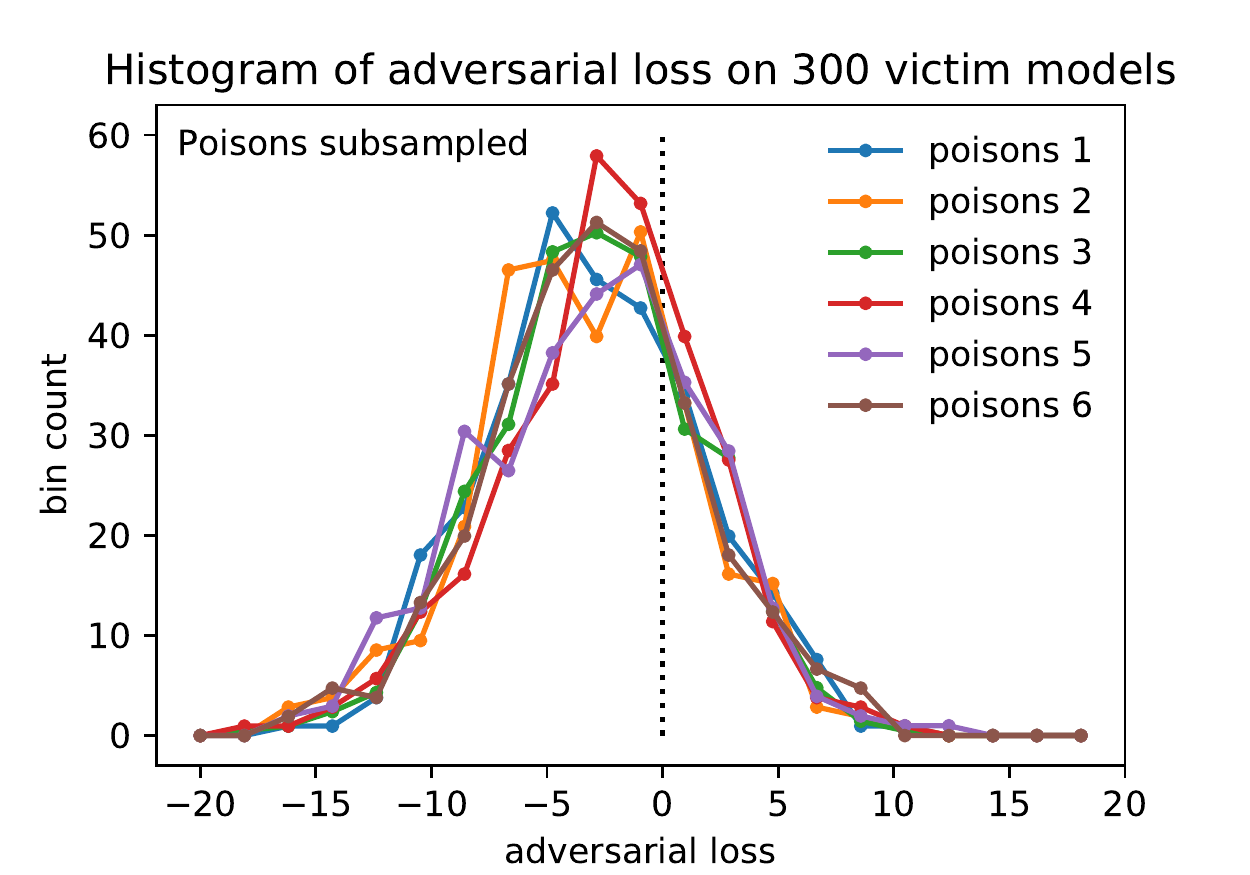}
    \caption{Poison crafting stability and subsampling. (Top) Histogram of adversarial loss from 300 different victim models. Each histogram represents a different set of 500 poison dogs crafted using different random seeds. (Bottom) Histogram of adversarial loss from 300 different victim models for a set of 500 poison dogs \textit{that are subsampled} from a set of 5000 poison dogs. The base IDs of the 500 subsampled poison dogs are identical to the 500 base IDs used in Figure \ref{fig:stability} (top).}
    \label{fig:stability}
\end{figure}

We craft 6 sets of poisons under the same settings (500 poison dogs, target bird with ID 5) with different random seeds and compare their victim evaluation results. Since there is already stochasticity in training a victim model even for the same set of poisons (see, e.g., \S\ref{sec:trainingcurves}), we train \textit{300} victim models on each set of poisons and plot a histogram of the resulting adversarial loss for each in Figure \ref{fig:stability} (top). The histograms overlap one another almost perfectly, indicating that the poison crafting process is generally pretty stable and the crafted poisons will have similar potencies from run to run. For this particular example, the adversarial loss distribution happens to center around zero, where the half on the left represent models that are successfully poisoned while the half on the right represent models that are not (a property of the Carlini and Wagner et al. (2017) loss function).

\section{Subsampling poisons from a larger set}
\label{sec:subsampling}

One practical question is whether poisons crafted together \textit{work together} to influence the victim training dynamics toward the desired behavior (i.e. lowering adversarial loss), or if each poison individually does its part in nudging the victim weights toward the desired behavior. Posed another way, if we subsample a larger poison set to the desired poison budget, would the resulting adversarial loss be the same as if we had directly crafted with the desired poison budget? This question is quite practical because in many cases the attacker cannot guarantee that the \textit{entire} poison set will be included into the victim dataset, even if some subset of it will likely trickle into the dataset.

We investigate the effect of subsampling poisons. We subsample a set of 500 poison dogs from a larger set of 5000 poison dogs. The 500 base IDs of the subset are identical to the base IDs used in Figure \ref{fig:stability} (top) for fair comparison. The poisons are crafted 6 times and the resulting adversarial loss histograms (each the result of 300 victim models) are shown in Figure \ref{fig:stability} (bottom).

First, notice that the histograms overlap in this case, again demonstrating the stability of the crafting process. Surprisingly, the histograms are more skewed toward negative adversarial loss than those in Figure \ref{fig:stability} (top), revealing that subsampling to the desired poison budget achieves better performance than crafting with the poison budget directly. This result is advantageous for the attacker because it relaxes the requirement that the \textit{entire} poison set must be included into the victim dataset without missing any. This result is also counter-intuitive as it suggests that  the \textit{direct} method crafting for a desired poison budget is inferior to the \textit{indirect} method of crafting for a larger budget and then taking a random subset of the poisons with the desired poison budget size. One possible explanation for this phenomenon may be that the higher dimensionality of a larger poison budget helps the optimization process find minima more quickly, similar to the way that the overparameterization of neural networks helps to speed up optimization \citep{sankararaman2019impact}.

Our experiments in the main paper, in Figures 3, 4, and 6, varied the poison budget by taking a different-sized subsets from a common set of 5000 poisons.

\section{Experiments on ImageNet-2k (Dogfish) dataset}
\label{sec:dogfish}

\begin{figure}
  \begin{center}
     \includegraphics[width=.6\textwidth]{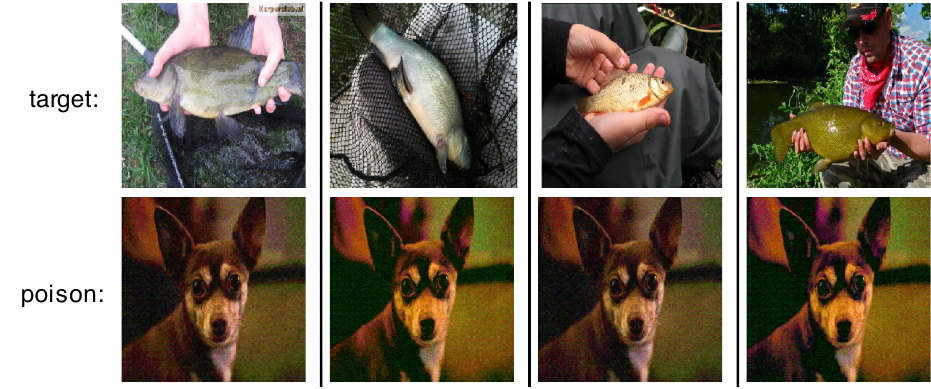}
     \caption{Examples of successful single poison dogs that cause the corresponding target fish above it to be misclassified as a dog on the Dogfish dataset.}
     \label{fig:dogfish}
  \end{center}
\end{figure}

\citet{koh2017understanding} and \citet{shafahi2018poison} also ran their poisoning experiments on the Dogfish dataset, which is a small subset of only two classes taken from ImageNet proper (see \citet{koh2017understanding} and associated code for more details), consisting of \~2000 examples. In their setups, a frozen, pretrained feature extractor was used and only the last, classification layer of the network was trained. Both acheived successful targeted poisoning with a single poison example, with \citet{shafahi2018poison} achieving 100\% success over many targets. To compare MetaPoison with prior work on this simple last-layer transfer learning task, we pretrained a 6-layer ConvNet (same as in the main paper, except with larger width and more strides to accommodate the larger images), and performed last-layer transfer learning on a poisoned Dogfish dataset, consisting of just one poison dog crafted under similar conditions as outlined in the main paper. Likewise, we found that we achieved 100\% success rate over the first 10 target fish images, while maintaining a validation accuracy of 82\% both before and after poisoning. Figure \ref{fig:dogfish} show some example poison dogs along with their corresponding target fish.

\section{Feature space visualizations for from-scratch training}
\label{sec:featvizapp}

\subsection{By epoch}
\begin{figure}
  \begin{center}
     \includegraphics[width=.5\textwidth]{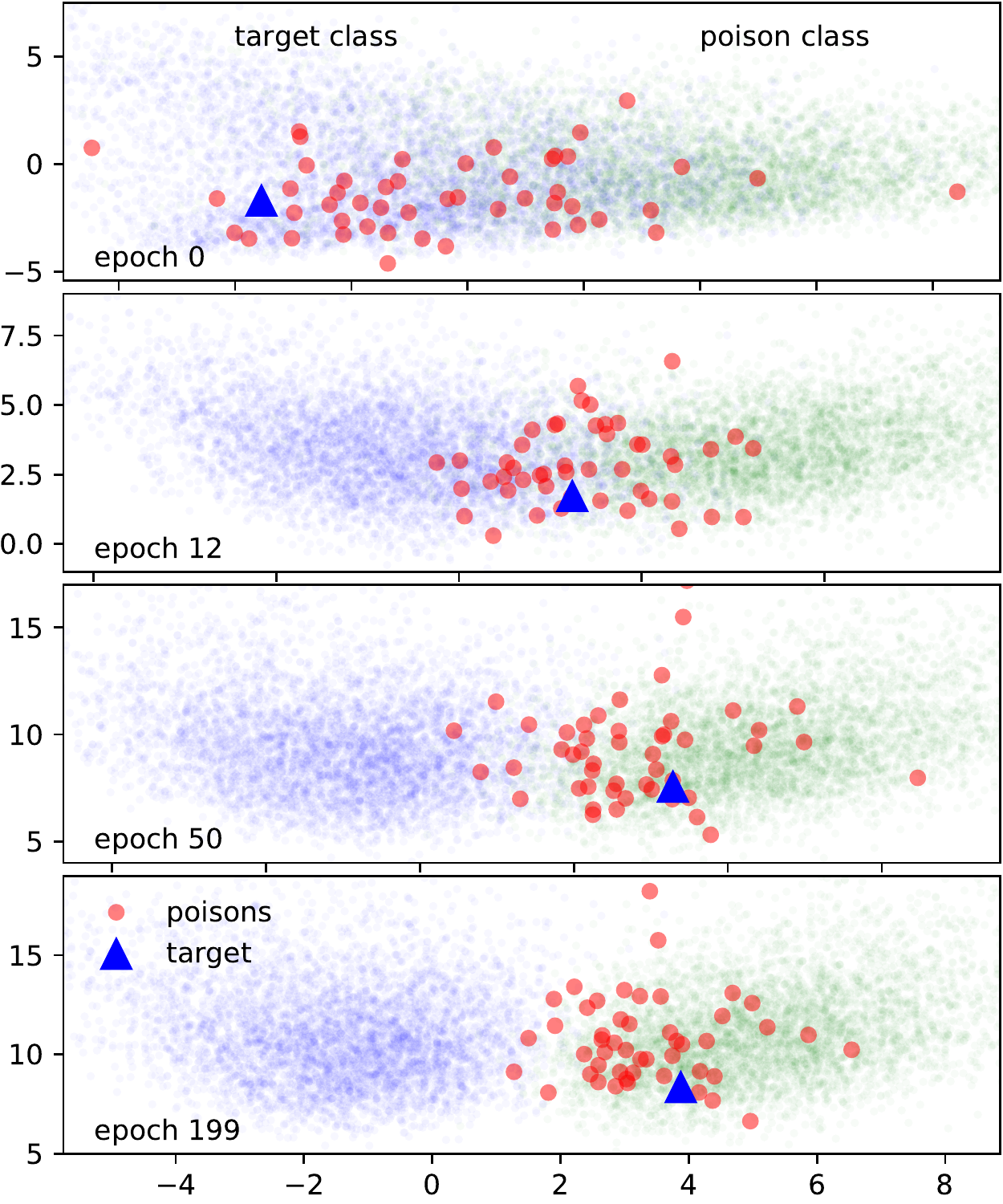}
     \caption{Penultimate layer visualization as a function of epoch for
a successful train-from-scratch attack of 50 poisons. The target
(blue triangle) is moved toward the poison distribution by the
crafted poisons.}
     \label{fig:featviz}
  \end{center}
\end{figure}
 Like in Figure 3 (bottom) We again gain clues to the poisoning mechanism through feature space visualization. We view the penultimate layer features at multiple epochs in the inset figure, showing a penultimate layer visualization as a function of epoch for a successful train-from-scratch attack of 50 poisons. The target (blue triangle) is moved toward the poison distribution by the crafted poisons. In epoch 0, the classes are not well separated, since the network is trained from scratch. As training evolves the earlier-layer feature extractors learn to separate the classes in the penultimate layer. They do not learn to separate the target instance, but they instead steadily usher the target from its own distribution to the poison class distribution as training progresses to epoch 199, impervious to the forces of class separation. In addition, the distribution of poisons seems to be biased toward the side of the target class. This suggests that the poisons adopt features similar to the specific target image to such an extent that the network no longer has the capacity to class-separate the target from the poisons. See the supp. material additional visualizations and insights.

\subsection{By layer}
\label{sec:layerviz}

\begin{figure}
    \centering
    \includegraphics[trim=0 2.3cm 0 2cm, clip, width=0.5\textwidth]{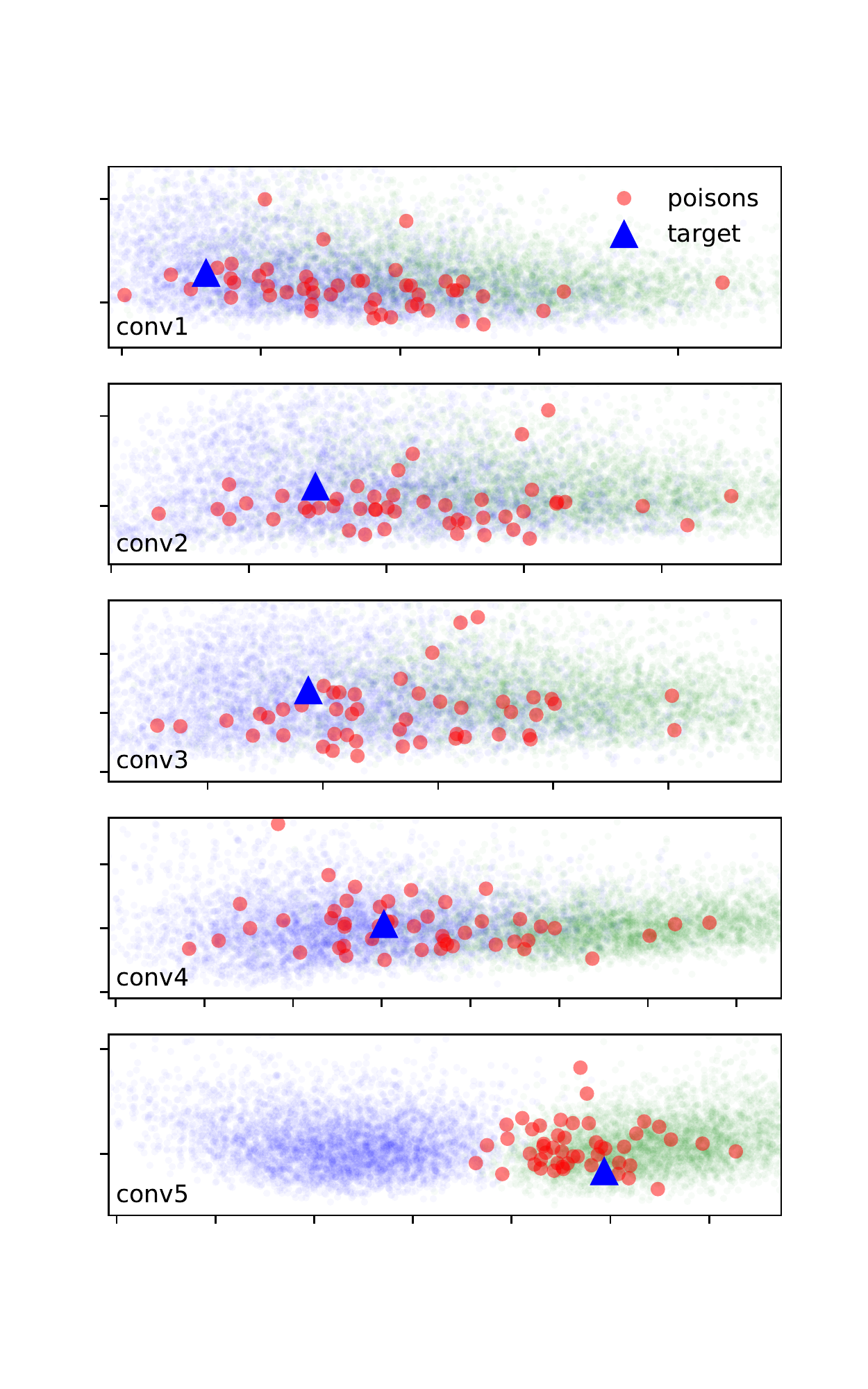}
    \caption{Feature visualization as a function of network layer in ConvNetBN for a successful attack of 50 poisons. Blue circles correspond to target class data, while green circles correspond to poison class data. The poisons (red circles) cluster in the target class cluster in the earlier layers. In the last layer, conv5, the poisons and target (blue triangle) move to the poison class cluster and the target is misclassified.}
    \label{fig:bylayer}
\end{figure}

Poisoned training data influences victim models to misclassify a specific target. While they are optimized explicitly to do this via a loss function, the \textit{mechanism} by which the poisons do this remains elusive. In addition to Figure \ref{fig:featviz}, we use feature visualization as a way to inspect what is happening inside the neural network during poisoning. Figure \ref{fig:featviz} showed the evolution of the features in the penultimate layer across multiple epochs of training. Here, in Figure \ref{fig:bylayer}, we visualize the evolution of the features as they propagate through the different layers of the trained (epoch 199) ConvNetBN victim network. The projection method used is the same as that in \S3.1.

Like in Figure \ref{fig:featviz}, the blue points on the left of each panel are data points in the target class, while the green points on the right are data points in the poison class. The target is denoted by the blue triangle and the poisons are denoted by red circles. The data in the two classes are initially poorly separated in the first layer (conv1) and become more separable with increasing depth. Interestingly, the poisons do not cluster with their clean counterparts until the last layer, preferring to reside in the target cluster regions. Even at conv5, the poisons reside closer to the target class distribution than does the centroid of the poison class. Like in \S\ref{sec:featvizapp}, this implies that they must adopt features similar to the target instance to ``rope in'' the target to the poison class side of the boundary. We see here especially that the features adopted by the the poisons are similar to the target at all levels of compositionality, from the basic edges and color patches in conv1 to the higher level semantic features in conv5. Additionally, MetaPoison is able to find ways to do this without the need to explicitly collide features. Finally, notice that neither the poisons nor target move to the poison class cluster until the final layer. This suggests that the poison perturbations are taking on the higher--rather than lower--level semantic features of the target. This behavior may also be a telltale signal that the target is compromised by data poisoning and could potentially be exploited for a future defense

\section{Further examples of data poisons}
\label{sec:morepoisons}

\begin{figure*}
    \centering
    \includegraphics[width=0.5\textwidth]{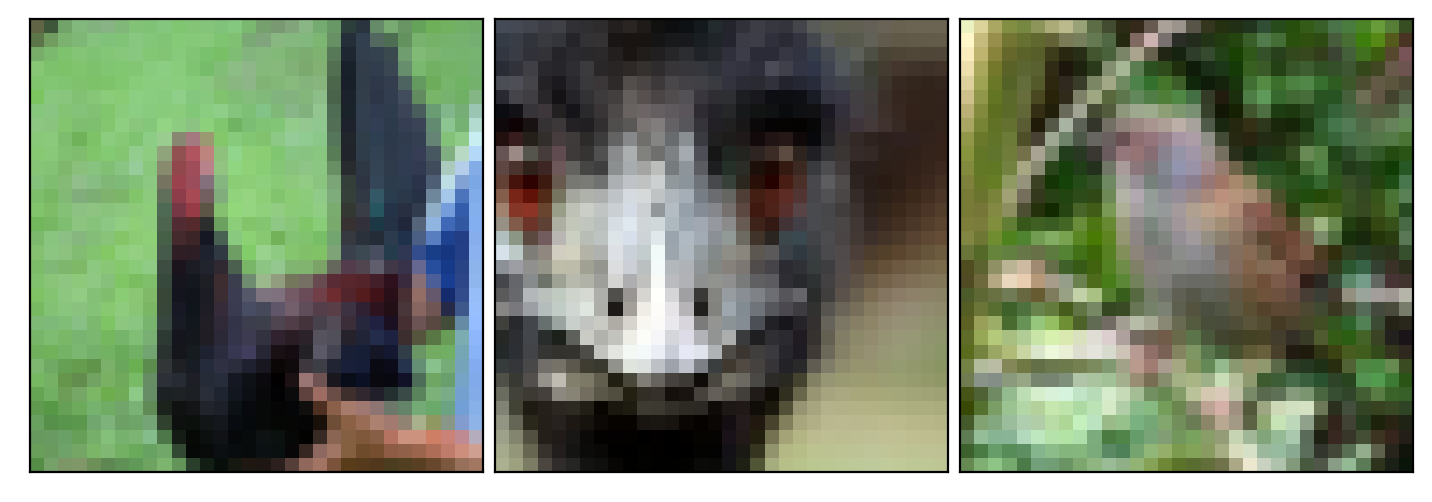}
    \includegraphics[width=0.8\textwidth]{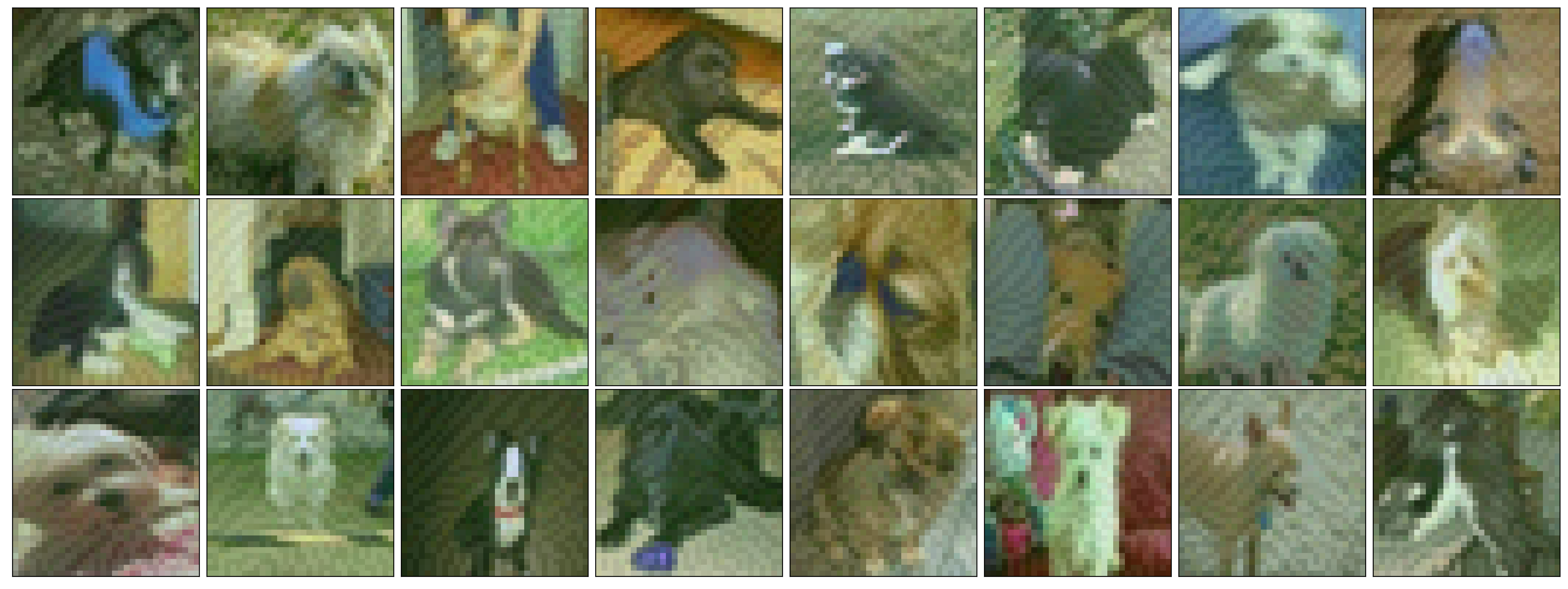}
    \includegraphics[width=0.8\textwidth]{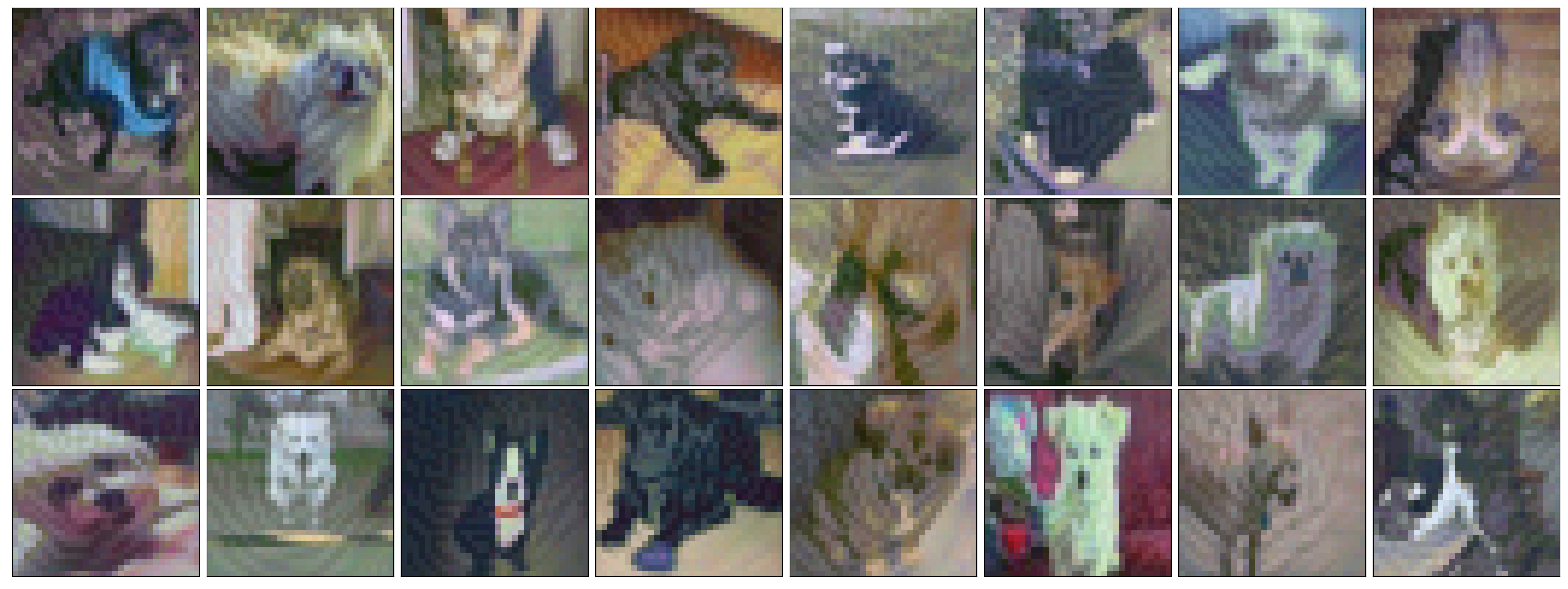}
    \includegraphics[width=0.8\textwidth]{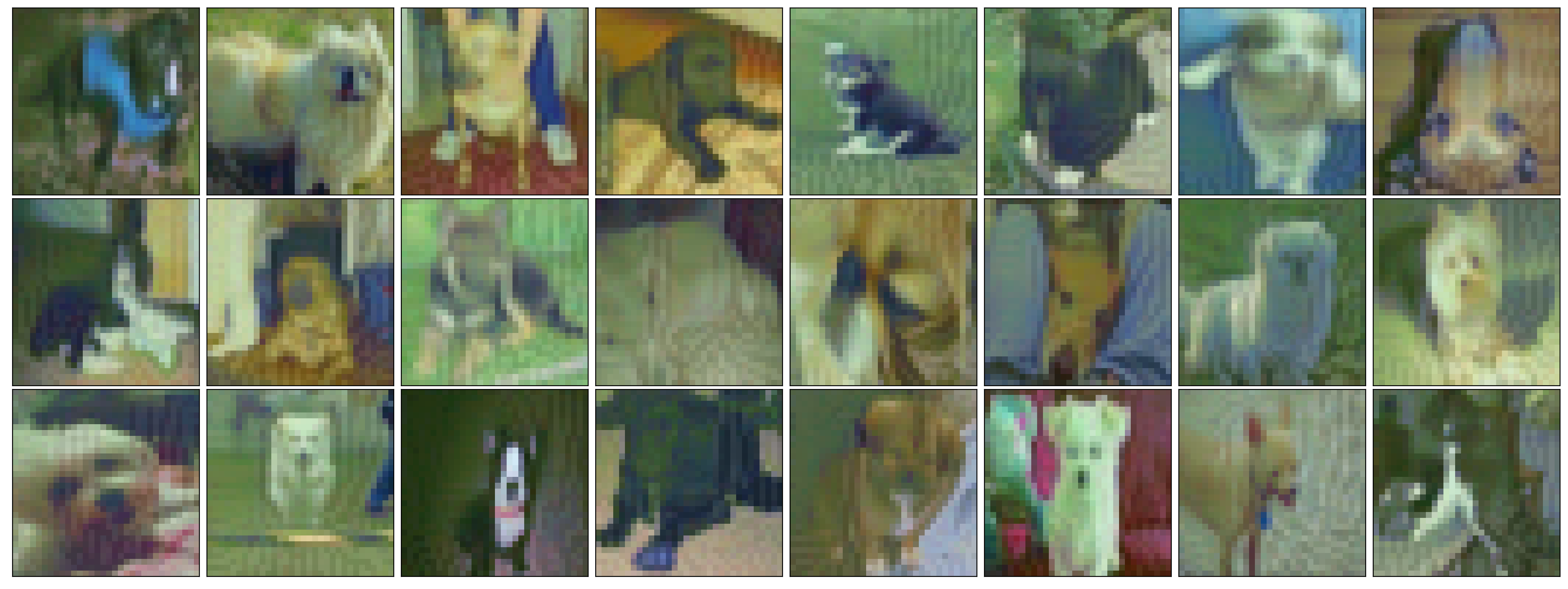}
    \caption{Poison dogs. These example poisons from top to bottom correspond to the targets from left to right, e.g. if poisons from the top 3x8 gallery are included in the training dataset, then the first bird is classified as a dog. Images shown are dogs 0-23 from the CIFAR training set and birds 0-2 from the CIFAR validation set.}
    \label{fig:lotsofpoisons2}
\end{figure*}

\begin{figure*}
    \centering
    \includegraphics[width=0.5\textwidth]{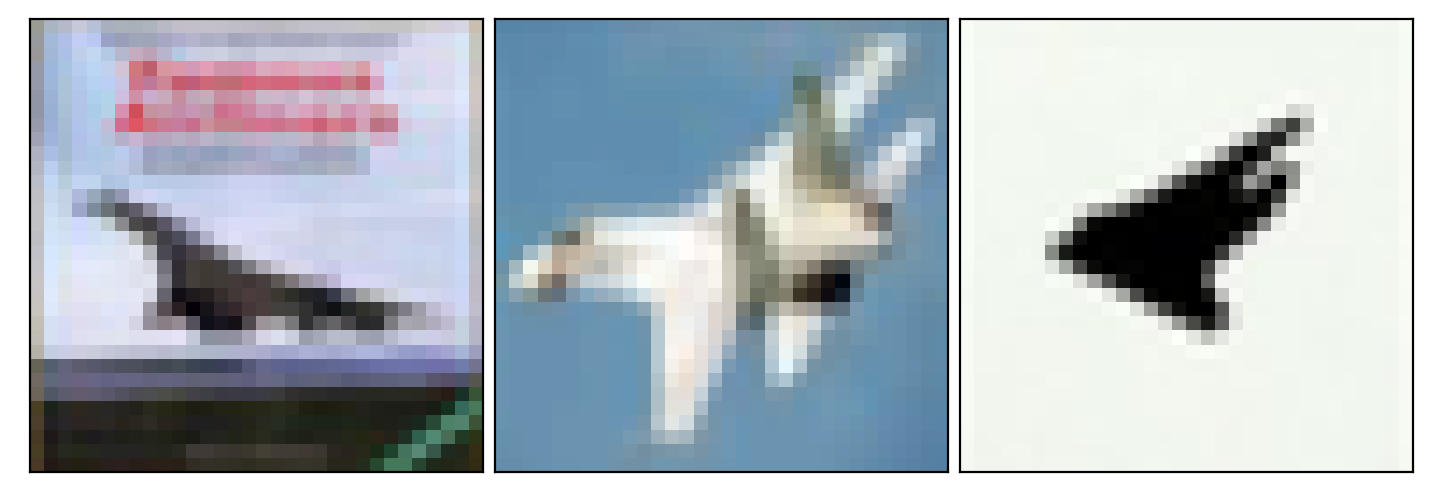}
    \includegraphics[width=0.8\textwidth]{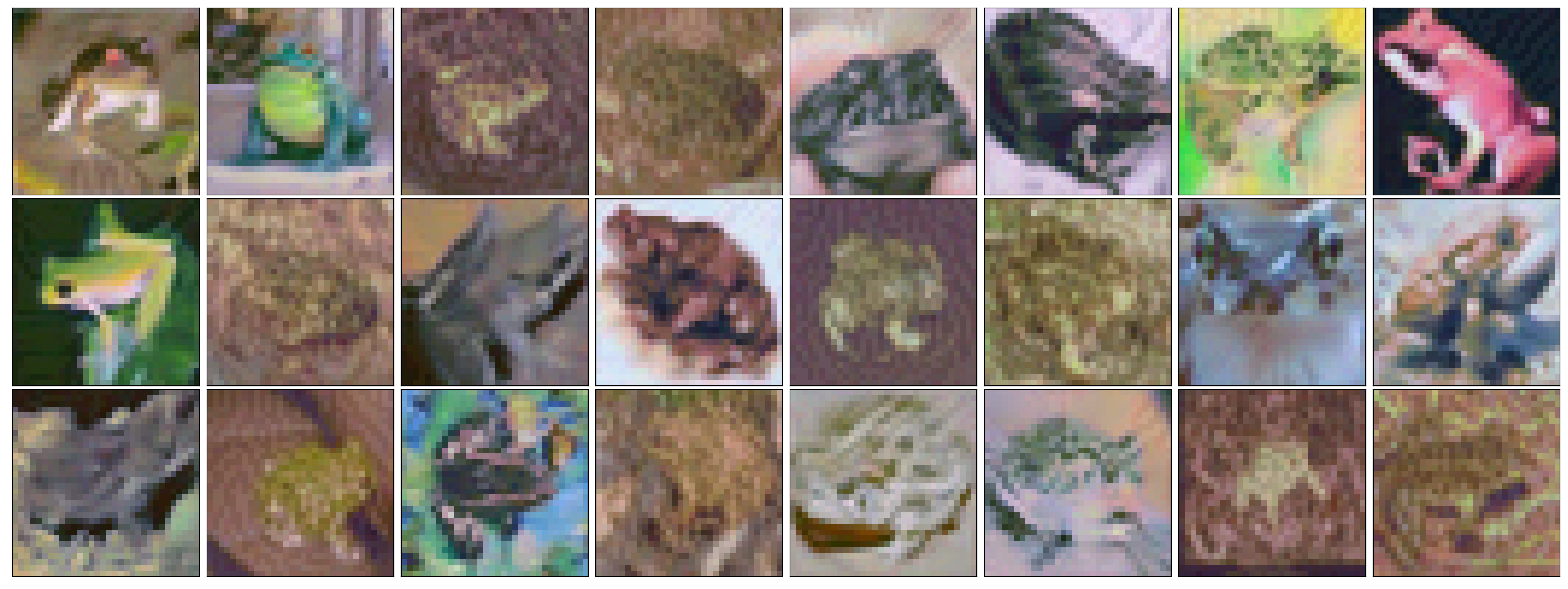}
    \includegraphics[width=0.8\textwidth]{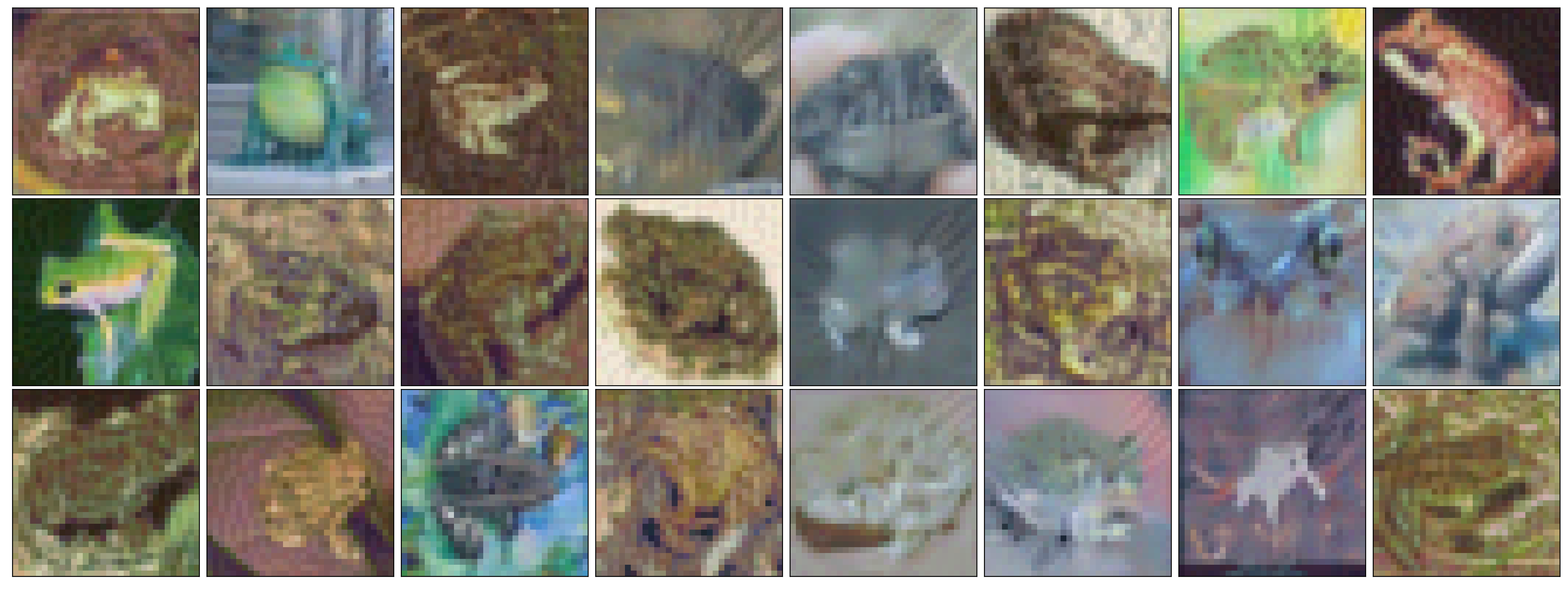}
    \includegraphics[width=0.8\textwidth]{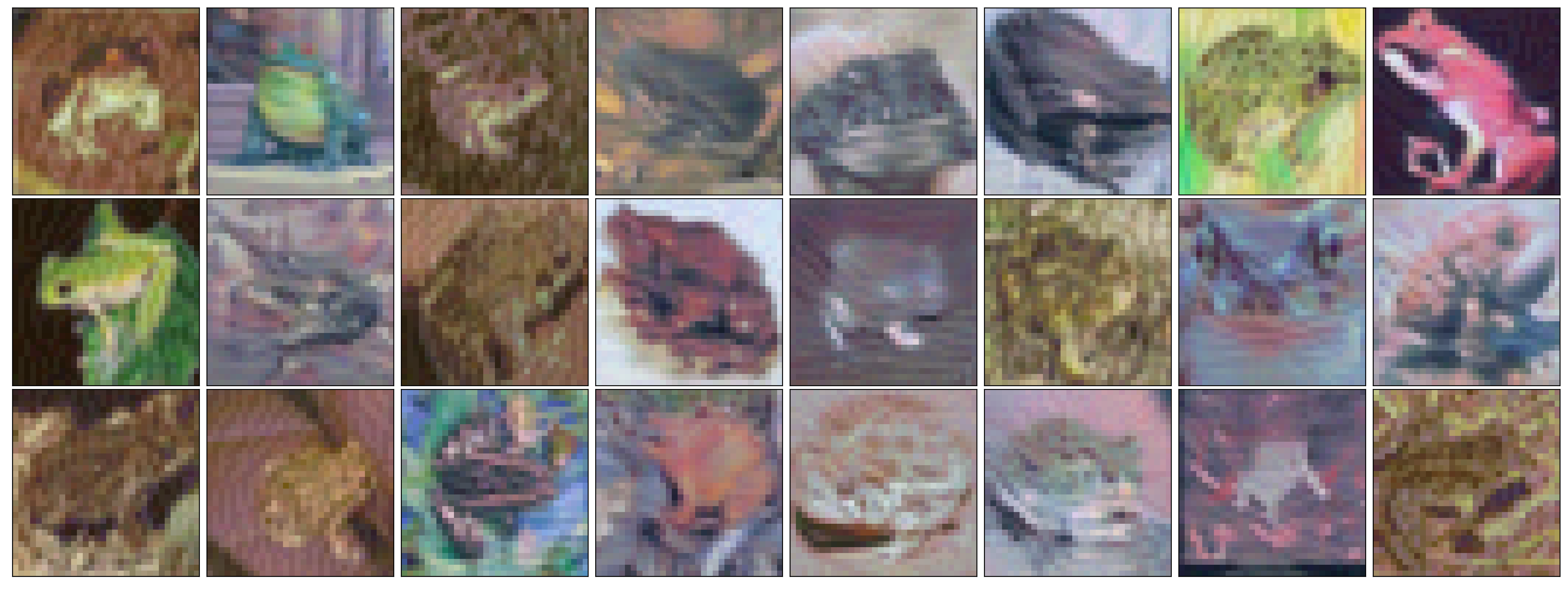}
    \caption{Poison frogs. These example poisons from top to bottom correspond to the targets from left to right, e.g. if poisons from the top 3x8 gallery are included in the training dataset, then the first airplane is classified as a frog. Images shown are frogs 0-23 from CIFAR training set and planes 0-2 from the CIFAR validation set.}
    \label{fig:lotsofpoisons1}
\end{figure*}

Figures \ref{fig:lotsofpoisons2} and \ref{fig:lotsofpoisons1} show more examples of the crafted data poisons in several galleries. Each gallery corresponds to a different target image (shown on top from left to right). These poisons are crafted with poison parameters $\epsilon=8$, $\epsilon_c=0.04$. We always show the first 24 poisons (in the default CIFAR order) for the first three target images taken in order from the CIFAR validation set.

\newpage

